\documentclass[preprint, review, 12pt, sort]{elsarticle}

\usepackage{graphicx} % Required for inserting images
\usepackage{amsmath}
\usepackage{subcaption}
\usepackage{bm}
\usepackage{amssymb}
\usepackage{placeins}
\usepackage[table]{xcolor}
\usepackage{array}
\usepackage[hidelinks]{hyperref}
\usepackage{fullpage}
\usepackage{booktabs}
\usepackage{multirow}

\newcommand{\eps}{\varepsilon}
\newcommand{\beps}{\bm{\varepsilon}}
\newcommand{\bsig}{\bm{\sigma}}
\newcommand{\bC}{\bm{C}}

\newcommand{\bz}{\bm{z}}

\newcommand{\candeps}{\hat{\beps}}

\newcommand{\bceps}{\overline{\beps}}

\newcommand{\trueeps}{{\beps}^*}

\newcommand{\impnet}{g^{\bC}_\phi}

\newcommand{\ddp}{\mathbin{:}}

\newcommand{\volavg}[1]{\langle #1 \rangle}

\begin{document}

\title{Thermodynamically-Informed Iterative Neural Operators for Heterogeneous Elastic Localization}
\author{Conlain Kelly, Surya R. Kalidindi}

\begin{abstract}
Engineering problems frequently require solution of governing equations with spatially-varying discontinuous coefficients. Even for linear elliptic problems, mapping large ensembles of coefficient fields to solutions can become a major computational bottleneck using traditional numerical solvers. Furthermore, machine learning methods such as neural operators struggle to fit these maps due to sharp transitions and high contrast in the coefficient fields and a scarcity of informative training data. In this work, we focus on a canonical problem in computational mechanics: prediction of local elastic deformation fields over heterogeneous material structures subjected to periodic boundary conditions. We construct a hybrid approximation for the coefficient-to-solution map using a Thermodynamically-informed Iterative Neural Operator (TherINO). Rather than using coefficient fields as direct inputs and iterating over a learned latent space, we employ thermodynamic encodings -- drawn from the constitutive equations -- and iterate over the solution space itself. Through an extensive series of case studies, we elucidate the advantages of these design choices in terms of efficiency, accuracy, and flexibility. We also analyze the model's stability and extrapolation properties on out-of-distribution coefficient fields and demonstrate an improved speed-accuracy tradeoff for predicting elastic quantities of interest. 
\end{abstract}

\maketitle

\section{Introduction}

The ability to model and predict the evolution of complex, heterogeneous spatial structures is central to materials design. Representing these systems computationally requires solving a set of governing equations, often partial differential equations (PDEs). In this work, we focus on parametric elliptic PDEs, which are characterized by different sets of coefficient fields and forcing terms. Consider a system of equations

\begin{equation}
    \mathcal{L}(a, u) = f 
\end{equation}
\noindent where $a(x)$ represents spatially-varying coefficients and $f(x)$ captures boundary conditions or forcing terms, $u(x)$ is the solution field, and $\mathcal{L}$ is a differential operator tacitly assumed to also consider derivatives of its arguments (such as $\nabla u$) as inputs. Oftentimes $f$ is fixed but $a$ itself is the output of some underlying generative process conditioned on some parameters (e.g., manufacturing settings) $\theta$. This process produces spatially heterogeneous structures; common scientific examples include material nano/microstructures \cite{robertson_efficient_2022,xu_establishing_2023}, geological systems \cite{dimitrakopoulos_high-order_2009, shekhar_fast_2023, lehmann_3d_2024}), and musculoskeletal or brain tissue \cite{zbontar_fastmri_2019, orozco_probabilistic_2024, bakhtiarydavijani_damage_2019}.

In general one can link sampled coefficient fields
\begin{equation}
a \sim P(a | \theta)
\end{equation}
\noindent with solutions $u$ by solving the governing equations, then postprocess $u$ to compute downstream quantities of interest $q$. Estimating forward and inverse uncertainties ($p(q | \theta)$ and $p(\theta | q)$, respectively) requires repeated evaluation of the \textit{coefficient-to-solution map}
\begin{equation}
    G: a \mapsto u \quad s.t. \quad \mathcal{L}(a, G(a)) = f.
\end{equation} 
\noindent In cases where only a few choices of $a$ or $f$ are required, a traditional numerical solver provides accurate and interpretable results \cite{atkinson_numerical_1967, liu_reproducing_1995, j_c_simo_computational_1998, brenner_mathematical_2008}. However, traditional solvers are typically optimized for accuracy and flexibility rather than throughput. 

Alternatively, Neural Operators are a class of Deep Learning models specialized for problems such as parameteric PDEs. Leveraging innate data parallelism and batched inference, neural operator surrogates ostensibly provide a high-throughput alternative to traditional solvers. Recently neural operators have shown great success in applications such as elastic wave propagation \cite{zhang_learning_2023, lehmann_3d_2024}, multiphase flow \cite{grady_model-parallel_2023, kovachki_neural_2021}, elastography \cite{tripura_wavelet_2023} and fluid dynamics \cite{marwah_deep_2023, lin_operator_2023} even on irregular domains \cite{li_fourier_2023}. However, training a neural operator which is robust to distribution shift is an ongoing challenge \cite{tim_roith_learning_2024, zhu_reliable_2023, lippe_pde-refiner_2023} -- one which substantially undercuts their utility in inverse design. A machine learning model's performance depends directly on the quality and diversity of training data available and is highest near the training set. However, for design and inversion problems one is likely to encounter coefficient fields not represented in the training distribution \cite{yin_solving_2023}.

To this end, a vast body of research has explored \textit{data-free} training, applying strong- or weak-form versions of the governing equations as loss functions on the neural operator's output \cite{anagnostopoulos_learning_2024,zeng_competitive_2022,raissi_physics-informed_2019, harandi_spectral-based_2024, song_physics-informed_2025}. Physics-based losses effectively shift the computational burden of data generation from a high-fidelity solver onto the training loop of the operator itself. However, this does not impact the internal architecture of the operator, and still specializes the model to certain subsets of the input/output spaces. Therefore we ask a complementary question: \textit{how can one design neural operators in a way which encourages scalability and transferability, regardless of the objective function?}

We draw on the rich literature of iterative numerical solvers \cite{saad_iterative_2003}: rather than computing a direct map from input to output, we wish to incrementally improve predictions until some stopping criterion is reached. In the context of Deep Learning, this idea is known as loop unrolling, implicit learning or Deep Equilibrium (DEQ) learning. Historically, these models have been primarily used mostly in inverse problems \cite{andrychowicz_learning_2016, putzky_invert_2019, monga_algorithm_2021}. However, a number of recent advances in training methodology \cite{winston_monotone_2020,fung_jfb_2022, anil_path_2022, gabor_positive_2024} have lead to increased use in applications such as image representation \cite{sitzmann_implicit_2020}, generation \cite{pokle_deep_2022}, and neural operators \cite{you_learning_2022, marwah_deep_2023}. A number of theoretical questions arise with the shift to implicit learning, in particular: what latent spaces and input representations are desirable for fast and accurate inference? 

% Why do we care about heterog localization
To explore these questions tractably, we focus on an archetypal parametric PDE: computing the deformations of heterogeneous material structures subjected to bulk loading conditions, hereafter referred to as the \textit{localization} problem. Since materials are inherently stochastic, heterogeneous, multiscale systems \cite{coleman_thermodynamics_1967, kalidindi_hierarchical_2015}, their internal structures are of central interest for both design and evaluation \cite{kalidindi_digital_2022}. In the following discussion, we restrict our attention to mesoscale elastic materials, so that different instantiations of coefficient fields (rank-4 symmetric elastic stiffness tensors) are referred to as microstructures. Likewise, we use pointwise stresses and strains as local solutions; while displacements or deformation mappings are the fundamental solution variable at a micromechanical level, predicting strains and stresses is sufficient for many homogenization problems \cite{hill_self-consistent_1965,kroner_statistical_1971,sab_fft-based_2024}.

In the following sections we introduce relevant background in deep learning and micromechanics, then develop and evaluate a class of neural operators which predict local strains and stresses over 3D heterogeneous elastic structures with known voxel-level constitutive information. To balance the above challenges, our method learns a Deep Equilibrium Model which iteratively refines strain predictions using thermodynamic encodings. A Fourier Neural Operator is used as the core filtering operation, and fixed-point iteration with Anderson Acceleration is used to solve the resulting system of equations. 

The contributions of this work are threefold: first, we present a solver for 3D heterogeneous elasticity, which has remained relatively untouched in literature. Since inelastic deformation mechanisms such as crystallographic slip are inherently three-dimensional, this work presents a bridge towards modeling much more complex and nonlinear phenomena with neural operators. Second, we present a thorough comparison of our method against several state-of-the-art iterative neural operator architectures in terms of efficiency, predictive accuracy, and inference stability. In particular, we find that inclusion of constitutive information inside the iteration loop tends to stabilize iterations, reduce numerical artifacts, and provide lower prediction error at the expense of slower internal convergence. Finally, we formulate a zero-shot extrapolation problem, wherein each model is tested on out-of-distribution structures which are more difficult (higher contrast) than its training set. In this setting we find that our model produces between two to five times smaller errors for the same parameter budget as other state-of-the-art methods. 

We now briefly introduce the notation used hereafter. All tensor quantities are represented using either boldface ($\bC$) or index notation ($C_{ijkl}$). Single and double dot products refer to single and double contractions along the innermost indices ($\bsig \cdot \mathbf{e}) = \sigma_{ij} e_j$, $\bC \ddp \beps = C_{ijkl} \beps_{kl}$). Scalar quantities are denoted without subscript or bold. For brevity, spatial dependence is shown in variable definitions but not in subsequent use, and all quantities are assumed spatially-varying unless otherwise stated. While continuous integrals are used for theoretical development, they computationally take the form of sums over voxel-wise quantities. Unless otherwise specified, all superscripts are used as indices rather than exponents. In order to efficiently store symmetric rank-2 and rank-4 tensors, we utilize Mandel notation \cite{mandel_generalisation_1965, manik_natural_2021} to convert these quantities to vectors of length $6$ and $6 \times 6$ matrices, respectively. Further detail on this process is presented in \ref{app:representation}.

\subsection{Parametric PDEs in materials}

A great amount of effort has been invested in experimentally exploring the connections between the process parameters used to synthesize or manufacture a material, its internal structure, and its thermal, electrical, and mechanical properties of interest. The combination of lab automation \cite{nist_materials_2014, kalidindi_data_2019, powell_snappy_2023} and data-driven methods \cite{fung_inverse_2021,pyzer-knapp_accelerating_2022,riebesell_can_2023, jin_mechanical_2023} offer a high-throughput exploration of materials spaces; however, bridging information between these spaces -- and across length-scales -- presents several challenges. The inherent complexity and stochasticity of material structures introduces a large amount of scatter in process-property mappings. Prior physical domain knowledge (e.g., atomistics,  constitutive models) and experimental data generally provide information regarding either the process-structure or structure-property cases. This introduces a major computational challenge -- estimating process-structure-property linkages requires repeated evaluation of both a generative (process $\rightarrow$ structure) model and a forward (structure $\rightarrow$ property) solver. The latter is exactly our parametric PDE above; since the mapping can be expensive to compute for a large ensemble of microstructures, it is often approximated or bypassed in construction of UQ or inverse design models. Common PDEs in material design involve (thermo-, visco-) elasticity \cite{eloh_development_2019}, continuum \cite{gurtin_mechanics_2010} and crystal plasticity \cite{roters_overview_2010, shanthraj_spectral_2019, zaiser_stochastic_2020}, and phase-field evolution \cite{brough_microstructure-based_2017, harrington_application_2022, hu_accelerating_2022}. 

On the coefficient side, a burgeoning literature of generative models for materials has been developed using statistical parametrizations \cite{jiao_modeling_2007, robertson_efficient_2022}, optimization methods \cite{fullwood_microstructure_2008, bhaduri_efficient_2021, kench_generating_2021}, Generative Adversarial Networks \cite{narikawa_generative_2022}, and diffusion maps \cite{lee_virtual_2021,robertson_local-global_2023, fernandez-zelaia_digital_2024, buzzy_statistically_2024}. Since the upstream field of Generative AI is rapidly growing and changing, we anticipate a continued growth in the availability of synthetic microstructure generators. The ability to rapidly and accurately evaluate microstructures, especially coming from a wide range of sources, further accentuates the need for a fast surrogate solver.

In summary, the challenge of materials design requires localization solvers which are both fast and accurate, perform stably and degrade gracefully outside their training distribution, and naturally accommodate the complexity of heterogeneous, stochastic material structures. While this work is directed specifically towards elastic deformation of mesoscale materials, the underlying system of equations is structurally quite similar to other engineering problems such as diffusion \cite{van_harten_exploiting_2024}, Darcy flow \cite{takamoto_pdebench_2024}, and electric conductivity \cite{willot_fourier-based_2014}. 

\subsection{Elastic Localization}
The periodic small-strain elastic localization problem is defined as the mapping from a microstructure instantiation $m(x)$ to its resultant strain field $\beps(x)$ when subjected to periodic loading conditions which impose an average strain field $\bceps$. Connecting displacement $u(x)$ and strain is the compatibility condition:
\begin{equation}
    \eps_{ij} = \frac{1}{2} \left(\frac{\partial} {\partial x_i} u_j + \frac{\partial} {\partial x_j} u_i \right) \iff \nabla \times (\nabla \times \beps) = 0. \\
\end{equation}

\noindent The heterogeneous microstructure is characterized by its (linear) elastic stiffness field $\bC(x)$, which relates strain $\beps(x)$, stress $\bsig(x)$, and strain energy density $w(x)$ via the constitutive relations

\begin{align}
    w(\beps, \bC) &= \beps \ddp \bC \ddp \beps \\
    \bsig(\beps, \bC) &= \frac{\partial w}{\partial \beps} = \bC \ddp \beps \\
    \intertext{Momentum balance results in the equilibrium condition}
    \nabla \cdot \bsig^T &= 0 
\end{align}
\noindent Periodic loading conditions are imposed by restricting the displacement field to the set $\mathcal{K}$ of displacement fields which produce periodic strain fields with average $\bceps$:

\begin{equation}
    \mathcal{K}_{\bceps} = \{u ~|~ \beps(u) = \bceps + \beps(v) \quad \text{for some periodic, continuous fluctuation field} ~ v \} \label{eq:constraint_set}
\end{equation}
\noindent This approach is commonly used in computational homogenization; for a deeper discussion regarding these conditions, the reader is directed to \cite{sab_fft-based_2024}.

While displacement is the primary micromechanical state variable, we can equivalently formulate the problem in terms of strains. This takes the form of predicting a periodic strain field $\beps$ given stiffness field $\bC$ and average strain $\bceps$:
\begin{align}
    \text{Map} \qquad \bC, ~ \bceps & \mapsto \beps \\
    s.t. \qquad \nabla \cdot (\bC \ddp \beps) &= 0 \\
    \nabla \times (\nabla \times \beps) &= 0 \\
    \langle \beps \rangle &= \bceps \\ 
    \beps \quad& \text{periodic}.
\end{align}
\noindent Notably, all dependence on microstructure enters via the microstructure-to-coefficient map $m \mapsto \bC$. For the case of $N$-phase composite materials, $m$ often takes the form of phase indicator functions; for a polycrystalline material, $m$ instead takes the form of Bunge-Euler angles \cite{bunge_4_1982} or Generalized Spherical Harmonic coefficients \cite{barrett_generalized_2019}. To avoid restricting our work to a single class of materials, we work directly with the coefficient field. 

While the governing equations themselves are linear and elliptic, the coefficient-to-solution mapping $\bC \mapsto \beps$ (with $\bceps$ fixed) is in fact nonlinear \cite{cohen_approximation_2015}. Moreover, microstructures are inherently discontinuous, discrete objects; for a composite material $\bC$ will be piecewise constant and discontinuous across grains or phase boundaries. This lack of regularity propagates into the solutions; while $u$ must be continuous, it is not differentiable across jumps in $\bC$, and thus $\beps$ is piecewise continuous (even before discretization). These complexities have spurred a rich literature of numerical solvers but hindered the application of learning-based methods. This problem has been explored extensively in continuum mechanics \cite{moulinec_numerical_1998, gurtin_mechanics_2010} as the linearization \cite{onsager_reciprocal_1931} of more complex phenomena such as finite-strain, inelastic, dissipative deformations. Therefore it acts as the outer loop for many numerical solvers \cite{j_c_simo_computational_1998, lebensohn_spectral_2020, krabbenhoft_basic_2002}.

% What is localization
\subsection{Traditional Localization Solvers}

While a vast number of discretization and approximation schemes have been explored for elastic localization, we primarily focus here on finite-element and iterative FFT schemes. However, a number of alternative approaches exist in numerical homogenization \cite{hill_self-consistent_1965, mori_average_1973, pham_strong-contrast_2003, liu_self-consistent_2016, huang_introduction_2023, bhattacharya_learning_2024}.

Finite Element Methods convert a set of governing equations into a weak form and compute approximate solutions in a finite-dimensional function space  \cite{germain_method_1973, brenner_mathematical_2008}. Sobolev approximation spaces provide control over the approximation's regularity, making them a natural fit for heterogeneous localization problems \cite{brezis_functional_2011}. These methods are extremely general in the physical geometries \cite{hughes_isogeometric_2005} and deformation mechanisms \cite{roters_overview_2010,mital_two-dimensional_2014,fischer_relating_2024, chakkour_developing_2024} they can model. However, this generality makes a general-purpose FEA solver poorly suited for many-query applications due to the cost of assembling and solving a global stiffness matrix for each microstructure in an ensemble. A number of remedies have been proposed, such as the element-by-element method \cite{wathen_analysis_1989}, and meshfree methods \cite{liu_reproducing_1995}. 

An alternative class of solution techniques -- often called FFT methods -- rely on a reformulation of the original governing equation into a more numerically favorable form. This is done by applying an operator splitting \cite{adomian_solving_1994, ryu_primer_2016} and Green's function solution \cite{green_essay_1828, eisler_introduction_1969} to convert the original system of differential equations into an implicit integral equation known commonly as the Lippmann-Schwinger (L-S) equation \cite{lippmann_variational_1950} or the Fredholm integral equation of the second kind \cite{atkinson_numerical_1997}. This reformulation gives rise to numerous homogenziation techniques \cite{kroner_statistical_1971, brown_solid_1955, torquato_effective_1997, pham_strong-contrast_2003}, as well as an array of iterative localization solvers. 

Notably, Moulinec and Suquet \cite{moulinec_numerical_1998}, demonstrated that for both elastic and inelastic deformations, the Lippmann-Schwinger equation is a contraction \cite{katok_introduction_1995, moulinec_convergence_2017} and can be solved efficiently via a matrix-free fixed-point iteration. Due to their use of Fast-Fourier transforms to approximate the Green's operator, these are often referred to as FFT solvers. A number of refinements have been proposed to the basic scheme, including different discretizations of the Green's operator and different update rules / search methods. Of particular note are the augmented-Lagrangian approach \cite{michel_computational_2001}, polarization- \cite{monchiet_polarization_2012} or displacement-based iterations \cite{lucarini_dbfft_2019}, adaptive iterations \cite{sab_fft-based_2024}, and extensions to finite-strain and irreversible deformations \cite{eisenlohr_spectral_2013, lebensohn_spectral_2020}. The central issues these methods face are ringing artifacts (sometimes called Gibb's oscillations) due to use of a truncated Fourier basis for the Green's operator \cite{brisard_periodic_2014, willot_fourier-based_2015, schneider_review_2021}, the restriction to a regular, uniform grid, and ill-conditioning caused by contrast in material properties. 

\subsection{Deep Learning for Localization}
% Other methods

While machine learning has traditionally explored maps between finite dimensional spaces (e.g., vectors of fixed length or sampled images), operator learning seeks mappings between infinite-dimensional function spaces \cite{li_neural_2020}. This is extremely useful in cases where a model needs to perform well across discretizations or resolutions, or when the resolution differs between available training data and desired evaluation scenarios. Recently a great number of operator learning methods have been proposed, including Deep Operator Networks (DeepONets) \cite{lu_learning_2021, goswami_physics-informed_2022}, Fourier Neural Operators (FNOs) \cite{li_fourier_2020, tran_active_2020, rashid_revealing_2023,qin_toward_2024}, Spectral Neural Operators \cite{fanaskov_spectral_2024}, Graph Kernel Networks \cite{li_neural_2020, huang_introduction_2023}, and Neural Kernel Operators \cite{lowery_kernel_2024}. Since this work employs FNOs, we briefly review them here; further review and comparison on other neural operators can be found in \cite{lu_comprehensive_2022, azizzadenesheli_neural_2024}.

FNOs are effectively convolutional neural networks which use band-limited spectral convolutions (via Discrete Fourier Transforms) \cite{francesca_spectral_2016} rather than spatial convolutions. Their ability to capture global patterns makes them them well-suited for modeling spatial phenomena with nonlocal or long-range interactions. Unlike meshfree neural operator architectures, they follow relatively standard implementation and training schemes. The original formulation can be applied on any size Cartesian input grid, and when trained on a fine enough resolution, their accuracy does not degrade with mesh refinement (unlike standard CNNs \cite{raonic_convolutional_2023}). They can also be extended to non-rectangular domains by means of coordinate embeddings \cite{gao_phygeonet_2021, li_fourier_2023}.

In context of heterogeneous materials, FNOs have been used to model quasistatic fracture in 2D \cite{rashid_revealing_2023}, elastic plane strain or stress in two-phase composites  \cite{gupta_accelerated_2023, bhattacharya_learning_2024,rezaei_finite_2024,li_fourier_2023, tran_active_2020}, and 2D brittle fracture and hyperelasticity \cite{you_learning_2022}. The majority of FNO research has focused on problems involving two spatial dimensions due to the dramatic increase in parameter counts from 2D to 3D. To address this, different factorizations and compressions have been explored to reduce the memory overhead while preserving predictive capacity \cite{tran_factorized_2023}.

While many canonical deep learning models are direct and feed forward in nature, iterative methods such as loop unrolling \cite{monga_algorithm_2021}, learned optimizations \cite{andrychowicz_learning_2016}, and weight-tying \cite{press_using_2017} have grown in popularity \cite{putzky_invert_2019}. Deep Equilibrium (DEQ) Models \cite{bai_deep_2019} solve a nonlinear system of equations during training and inference to emulate an infinite-depth network by leveraging the Implicit Function Theorem \cite{fung_jfb_2022} to estimate gradients for backpropagation. A number of variations have been suggested to enforce contractivity \cite{winston_monotone_2020, gabor_positive_2024} and path-independence \cite{anil_path_2022}, as well as expressive capacity \cite{bai_multiscale_2020}. Essentially, DEQ models allow the user to trade training and inference time for higher expressive capacity and accuracy, controlled by the number of iterations and choice of nonlinear solver. 

\subsection{Comparison with Existing Work} \label{subsec:compare}
While elasticity and plasticity have been extensively explored in 1D and 2D \cite{bhattacharya_learning_2024,rezaei_finite_2024,li_fourier_2023, tran_active_2020}, there is a dearth of works which use machine learning to predict full 3D solutions over heterogeneous microstructures \cite{yang_establishing_2019, kelly_rln_elasticity_2020}. Among the myriad works in physics-based, operator, and implicit techniques, the closest to our proposed model are Implicit Fourier Neural Operators (IFNOs) by You et al. \cite{you_learning_2022} and Deep Equilibrium FNOs by Marwah, et al. \cite{marwah_deep_2023}. 

IFNOs use a fixed-point iteration inside the network's forward pass (similar to DEQ methods), but only iterate on the FNO's internal latent space. That is, the encoding and decoding operations are only used once. The motivation in that work was primarily to allow stable training of deeper FNO models without vanishing gradient issues or an exploding parameter count. Similarly, Marwah et al. employed a deep-equilibrium FNO to model porous composites (Darcy Flow) as well as steady-state incompressible flow in 2D \cite{marwah_deep_2023}. Rather than using fixed-point iteration and backpropagating through iterations, they used accelerated schemes (e.g., Anderson mixing, Broyden's method) and the Implicit Function Theorem to estimate gradients. However, they also used the same (encoder, implicit filter, decoder) architecture as You, et al. 

We first cast these variants in a shared notational framework. Let $B_n$ represent the ``body'' of an FNO (itself composed of $n$ Fourier layers $L$, with $\mathcal{F}$ representing the FFT operation):

\begin{align}
    L^k(v) &= \sigma(W^k v + \mathcal{F}^{-1} [R^k \mathcal{F}[v]]) \\
    \mathcal{B}_n &= L^{n} \circ L^{n-1} \circ \ldots \circ L^{1}
\end{align}

We use $P$ and $Q$ to denote encoder and decoder layers, respectively, which are often typically implemented via channel-wise linear transforms or MLPs. Using this notation a traditional FNO approximates the mapping $a \mapsto u$ (where $\hat{u}$ represents the predicted solution) as:

\begin{align}
    \hat{u} &= Q \circ B_n \circ P \circ a \\
    \intertext{The iterative versions described above differ in the choice of latent space and solution scheme, leading to the following variations:}
\begin{split}
    h^0 &= \mathcal{P} \circ a \\
    h^{k+1} &= h^k + \frac{1}{m} \mathcal{B}_n  (h^k) \\
    \hat{u} &= Q \circ h^m 
\end{split} \qquad \qquad\qquad \qquad \text{(IFNO, $m$ iterations)} \\
\intertext{and}
\begin{split}
    h^0 &= \mathcal{P} \circ a \\
    h^* &= h^0 + \mathcal{B}_n (h^*) \\
    \hat{u} &= Q \circ h^*
\end{split} \qquad \qquad\qquad \qquad \text{(FNO-DEQ)}
\end{align}

Note that both architectures only change the internal structure of the filtering operations, but leave the lifting and projection unchanged. Furthermore, the iterative models work over a \textit{learned} latent space which must simultaneously represent the input coefficients and the iterative candidate solutions, then be projected out into the solution space. While one could ostensibly apply knowledge of the underlying system of equations and thermodynamic structure into the internal architecture, it is not clear how this would work on a learned latent space.

In contrast, we propose iterating directly over the solution field $u$ and instead injecting $a$ into the iterative process via \textit{thermodynamic} embeddings $a, u \mapsto z$. While explained further below, our approach can be cast in the above notation as:

\begin{equation}
\begin{split}
    u^0 &= 0 \\
    \hat{u} &= \mathcal{Q} \circ \mathcal{B}_n \circ \mathcal{P} \circ z(a, \hat{u}) \\
\end{split} \qquad \qquad\qquad\qquad \qquad \text{(TherINO)}
\label{eq:therino_first}
\end{equation}

Another work in a similar vein is the Hybrid Iterative Numerical Transferable Solver (HINTS) framework \cite{zhang_hybrid_2022, zhang_blending_2024} developed by Zhang, et al. Rather than learn an implicit operator, HINTS interleaves calls to a standard relaxation solver (e.g., Jacobi or Gauss-Seidel iterations) with evaluations of a feedforward DeepONet to solve linear differential equations. In this regime the network effectively acts as a nonlinear preconditioner, taking in residuals and returning improved solution estimates. In this case the iterative space is exactly the solution space itself, and the network need only approximate the residual-to-error mapping (e.g., error in $F(a, u)$ vs error in $u$). While the HINTS paper does explore random coefficient fields, they focus exclusively on smooth and slowly-changing coefficient fields in one spatial dimension. We believe that a combination of these approaches may be valuable: alternating relaxation iterations (such as steps of an FFT solver) with the neural operator could be interpreted either as preconditioning the neural operator with a relaxation solver, or vice-versa. Since this raises a number of practical and theoretical questions, we leave it to future work.  

A number of works have also used neural networks as hybrid \textit{time integrators} by interleaving timesteps between neural and traditional solvers \cite{hu_accelerating_2022, oommen_rethinking_2024}. In this work, however, we focus on quasi-static problems. Likewise, deep learning has been used to construct initial guesses for Newton iteration \cite{aghili_accelerating_2024, odot_deepphysics_2022} and select tuning parameters for multigrid preconditioners \cite{caldana_deep_2024}. 

There are a number of other ``thermodynamics-informed'' learning techniques which enforce a thermodynamically-admissible \cite{horstemeyer_historical_2010, gurtin_mechanics_2010} structure upon a neural network \cite{masi_thermodynamics-based_2021,he_thermodynamically_2022, huang_variational_2022, hernandez_thermodynamics-informed_2022}. These works are orthogonal to ours; rather than learn flows within a constraint space, we use thermodynamics to inform the network's trajectory in an unconstrained space. In applications with dissipation and entropic constraints, a combined approach will likely be necessary (i.e., enforcing non-negative entropy production \textit{architecturally}). Likewise, the field of \textit{physics-informed} learning generally applies governing equations as loss functions, rather than architectures. Finally, \textit{derivative-informed} learning uses partial knowledge of derivatives of the coefficient-to-solution map ($\nabla_a G$) as an additional loss term \cite{oleary-roseberry_derivative-informed_2024}. All losses used in this work are data losses, but given access to an efficient (batched GPU) estimator of FEA weak-form losses, one could easily use that as an additional regularization.

\FloatBarrier\clearpage\section{Methodology}
In this section we switch notation to match that of elastic localization. While our model is built with elasticity in mind, it could be seamlessly translated by substituting $a$ for $\bC$ and $u$ for $\beps$. In this notation, Eq. \eqref{eq:therino_first} can be restated as approximating the coefficient-to-solution map $\bC \mapsto \beps$ with an implicit neural operator which iterates on the \textit{solution field itself}:
\begin{align}
    \hat{\beps} &= \impnet(\hat{\beps}) 
\end{align}

\noindent using a function approximator $g$ with parameters $\phi$. The architecture of $g$ can be any differentiable operator, including a call to a traditional numerical solver. 

Crucially, this formulation removes the requirement that $\bC$ be a direct input to any learned layers; instead we lift an $(\beps, \bC)$ pair using a pre-defined thermodynamic encoding $\bz$ and subsequently apply a standard neural network $g_\phi: \bz \mapsto \beps'$ to predict a new candidate strain field $\beps'$. The nonlinear system of equations to solve at inference time thus takes the form:

\begin{equation}
    \candeps = g_\phi(\bz(\candeps, \bC)) := \impnet(\candeps) \label{eq:NLsys}.
\end{equation}

\noindent That is, the neural operator $g_\phi$ is oblivious to the coefficients $\bC$ \textit{except for the thermodynamic encodings} $z$. This has significant theoretical benefits: even if the local coefficient fields are novel (out-of-distribution) to the neural operator, the local strains and stress may still familiar. For brevity we reuse the character $g$ between the update network $g_\phi: \bz \mapsto \beps'$ and the fixed-point operator $\impnet: \beps \mapsto \beps'$. 

Our proposed method can be conceptually partitioned as follows: 
\begin{enumerate}
    \item A lifting operation $\bz(\beps, \bC)$ which encodes candidate strain fields and microstructure into features for the update network using constitutive informaiton
    \item An update network $g_\phi: \bz \mapsto \beps'$ which produces an improved candidate strain field
    \item A root-finding method for solving the nonlinear system of equations Eq. \eqref{eq:NLsys}
\end{enumerate} 
\noindent In the remainder of this section we will discuss choices for these three operations, as well the data and training pipelines used to calibrate our model.

\subsection{Thermodynamic encodings}
Central to deep learning is the question of representation: how do we provide information to a model such that it will learn meaningful and useful mappings? In general, characterizing a universal latent space for microstructures -- and collecting enough data to effectively explore it -- remains an ongoing challenge \cite{robertson_local-global_2023, robertson_micro2d_2024, roding_predicting_2020}.

However, for a given $(\beps, \bC)$ pair the constitutive equations provide several convenient local thermodynamic descriptors such as stress $\bsig(\beps, \bC)$ and strain-energy density $w(\beps, \bC)$. In the linear-elastic case, these take the forms:
\begin{align}
    \bsig(\beps, \bC) &= \bC \ddp \beps \\
    w(\beps, \bC) &= \beps \ddp \bC \ddp \beps.
\end{align}

We use these thermodynamic descriptors to encode a candidate strain field $\beps^k$ and stiffness $\bC$ into a latent vector field $\bz^k$:

\begin{equation}
    \bz^k = \begin{bmatrix}
        \beps^k \\
        \bsig(\beps^k, \bC) \\
        w(\beps^k, \bC) \\
        \vdots
    \end{bmatrix}
\end{equation}
\noindent This operation can be used for any material class $m \mapsto \bC$ since it only uses the stiffness field (and not $m$ directly). As an alternative one could conceivably flatten $\bC$ into a vector representation and remove redundant entries; however, in the fully-anisotropic case this is a 21-channel vector field. This dimensionality poses a data curation challenge -- in order for the model to generalize well, its training data must contain samples which activate all 21 components. 

In contrast, by using thermodynamic encodings the effective input space is simply the set of strain fields and downstream products (such as stress and energy). We argue -- and demonstrate in the second and third case study -- that this modification drastically improves the model's data efficiency for cross-task generalization. Rather than activating all 21 independent elements of the elastic stiffness tensor, one must simply cover the desired range of achievable strain and stress fields. However, a network trained on a specific class of structures or boundary conditions is still not expected to perform as well when applied to different structure classes and loading conditions. 

\subsection{Architecture for update rule}
For the backbone $g_\phi$, one could conceivably use any voxel-to-voxel model. The primary desirable properties are:

\begin{itemize}
    \item Fast and memory-light inference during training and testing
    \item Ability to capture global patterns and handle roughness in output field
    \item Preservation of translation-equivariance and periodicity
    \item Resolution-invariance
\end{itemize}

\noindent In past work \cite{kelly_recurrent_2021} we used a V-Net architecture which operated strictly in real space; this broke translation-equivariance due to use of spatial down- and up-sampling. Standard convolutional neural networks have also been shown to perform poorly in super-resolution tasks due to their function-space inconsistencies \cite{raonic_convolutional_2023}. Among the myriad options for the filtering block in our model, we chose a Fourier Neural Operator due to ease of implementation, computational efficiency, and their relative popularity within literature, especially implicit-learning research \cite{you_learning_2022, marwah_deep_2023}. However, we do not make any statements about the relative performance of FNOs relative to other Neural Operator architectures in this work. Note that for our model we include the lifting and projection layers ($\mathcal{P}$ and $\mathcal{Q}$) \textit{within} $g_\phi$, so that the internal filtering can still explore a higher-dimensional representation space.

\subsection{Solving the nonlinear system}  \label{subsec:update_scaling}
Once the lifting and update rules are chosen, all that remains is solving the implicit equation \eqref{eq:NLsys}. The simplest approach is a depth-limited fixed-point iteration which stops after $N$ steps:

\begin{align}
    \beps^{k+1} &= g_\phi(\bz(\beps^k, \bC)) \\
    \candeps &= \beps^{N} .
\end{align}
\noindent This is effectively the IFNO approach, and can be treated as kind of discretized neural ODE with a forward Euler time-stepping \cite{you_learning_2022}. However, back-propagating directly through these iterations requires checkpointing of intermediate activations, limiting the depth which can be used at training time due to memory pressure. Moreover, this approach learns terminal solutions, not fixed points. This means that the model must always be evaluated for the same number of iterations, prohibiting the usage of early stoppage or longer rollouts at inference time.  

\subsection{Training process} \label{subsec:losses}

Given a choice of encoding, update operator, and iteration method, we calibrate our model by fitting it to a precomputed set of (stiffness, strain) pairs obtained using a Finite-Element solver. The spatial heterogeneity makes it challenging to choose a loss function which balances different regions of the microstructures: penalizing strain error alone can lead to poor stress predictions (since stiffer regions can have low strain but high stress), and vice-versa. Denoting the squared L2 function norm of a rank-2 tensor field $\beps$ with (symmetric, positive-definite rank-4 tensor field) weighting $\mathbf{A}(x)$ as 

\begin{equation}
    \| \beps \|^2_{\mathbf{A}} = \int \beps \ddp \mathbf{A} \ddp \beps  ~dV
\end{equation} we define strain-, energy-, and stress-based function norms for the misfit between $\trueeps$ and $\candeps$

\begin{align}
    L^{strain}(\trueeps, \candeps) &= \|\trueeps - \candeps \|^2_{\mathbf{I}}\label{eq:loss_strain}\\
    L^{energy}(\trueeps, \candeps) &= \|\trueeps - \candeps \|^2_{\mathbf{\bC}}\\
    L^{stress}(\trueeps, \candeps) &= \|\trueeps - \candeps \|^2_{\mathbf{\bC \ddp \bC}}\label{eq:loss_stress}
\end{align}

Note that all quantities used here are assumed to be adimensionalized as in \ref{app:representation}, such that the three losses are all unitless and roughly the same magnitude. Penalizing the energy of the strain error seems like the most elegant choice, but in fact lead to poor model convergence and degraded quality of stress estimates. We tested minimizing various combinations of losses, and found that simply summing the strain and stress losses led to the most robust and stable training. Empirically we found that the more powerful the model, the less sensitive it was to different weightings: for example training the simplest model (feedforward FNO) with only stress losses produced incredibly poor local strain estimates. 

Interestingly, this summation can itself be viewed as a matrix-weighted norm on the strain error:
\begin{equation}
    L^{total}(\trueeps, \candeps) = \|\trueeps - \candeps \|^2_{\mathbf{I} + \bC \ddp \bC}
\end{equation}
\noindent We also tried penalizing estimators for stress equilibrium or compatibility -- such as finite differences or Fourier differentiation -- but found that noise due to estimation and discretization outweighed any possible regularization benefits. Notably, since the underlying system of equations is linear, these estimators generally act as \textit{additional weighting terms} in the norm above. We hypothesize that addition of these terms would therefore be more helpful when the underlying system is nonlinear in the solution variable. Additionally, using a continuously-evaluable (e.g., DeepONet) architecture would potentially allow for more robust estimation of these derivatives or a weak-form loss against a set of test functions. 

\section{Experiments}

In this section we benchmark our proposed methodology against existing methods through a series of case studies. In particular, we address the following questions:

\begin{itemize}
    \item What are the advantages and disadvantages of using iterative/equilibrium neural operators compared to feed-forward neural operators?
    \item Are thermodynamic encodings useful representations of microstructure for inference?
    \item How accurate and stable are the fixed points learned by equilibrium neural operators?
    \item Which model form provides the best error-accuracy tradeoff?
\end{itemize}

To facilitate this comparison, we including the following models in our comparison: a standard feed-forward Fourier Neural Operator (\textbf{FNO}); a deeper and wider FNO (\textbf{Big FNO}); an implicit FNO (\textbf{IFNO}); a deep-equilibrium FNO using a learned latent space (\textbf{FNO-DEQ}); a modified FNO-DEQ which iterates over strain space but does not use thermodynamic encodings (\textbf{Mod. F-D}); and finally a Thermodynamically-informed Iterative Neural Operator (\textbf{TherINO}). The Big FNO model contains 151 million tunable parameters, and all others have 18.9 million parameters. All neural-operator models have identical hyperparameters and FNO architectures, but vary in the usage and ordering of FNO components. This allows a fair comparison between different model forms; however, it means that our implementations, hyperparameters, and model sizes differ slightly from those used in each model's original publication. While we observed some variation in model performance with changing architecture and depth, the relative performances remained unchanged. The enormous design space makes any kind of exhaustive hyperparameter search infeasible, but a limited ablation study is presented in \ref{app:arch}.  

To compare models quantitatively we use normalized (unsquared) L2 errors in local strain and stress fields. We also calculate L1 errors in equivalent (also known as von Mises) stress fields as well as absolute errors in homogenized effective stiffnesses to evaluate features that were not directly prioritized during training. Equivalent stresses were chosen because of their connections to yield and failure in plastic deformation \cite{dunne_introduction_2005}. Details for calculating all of these quantities are found in \ref{app:errors}. Qualitatively, we compare models by visualizing high-error slices of their outputs and errors within a representative microstructure. 

To develop and validate our model, a total of $10,000$ 2-phase microstructures were generated and split $40\% / 20 \% / 40\%$ into train/validation/test sets respectively. The test set was only used to generate final evaluation results for publication. Each microstructure was generated using the level sets of a Gaussian Random Field and simulated using the Moose Finite Element framework. In all cases a 0.1\% strain was applied (e.g. $(\bceps \ddp \bceps)^{1/2} = 0.001$). Each phase was assumed isotropic with equal Poisson ratios $\nu_1 = \nu_2 = 0.3$. The Young's modulus of the ``soft'' phase was fixed to $E_1 = 120$ (arbitrary units), and the ``hard phase'' had stiffness scaled by a contrast ratio $\kappa$ such that $E_2 = \kappa E_1$.  The first case study focuses on the fixed-contrast, fixed-loading scenario, the second on varying loading directions, and the third on variation in contrast; the same microstructure splits were used across all case studies. All models were trained and evaluated on an Nvida A100 40GB GPU. 

\FloatBarrier

\subsection{Fixed contrast and boundary conditions}

\begin{figure}[h]
    \centering
    \begin{subfigure}[b]{0.49\textwidth}
    \centering
    \includegraphics[width=\textwidth]{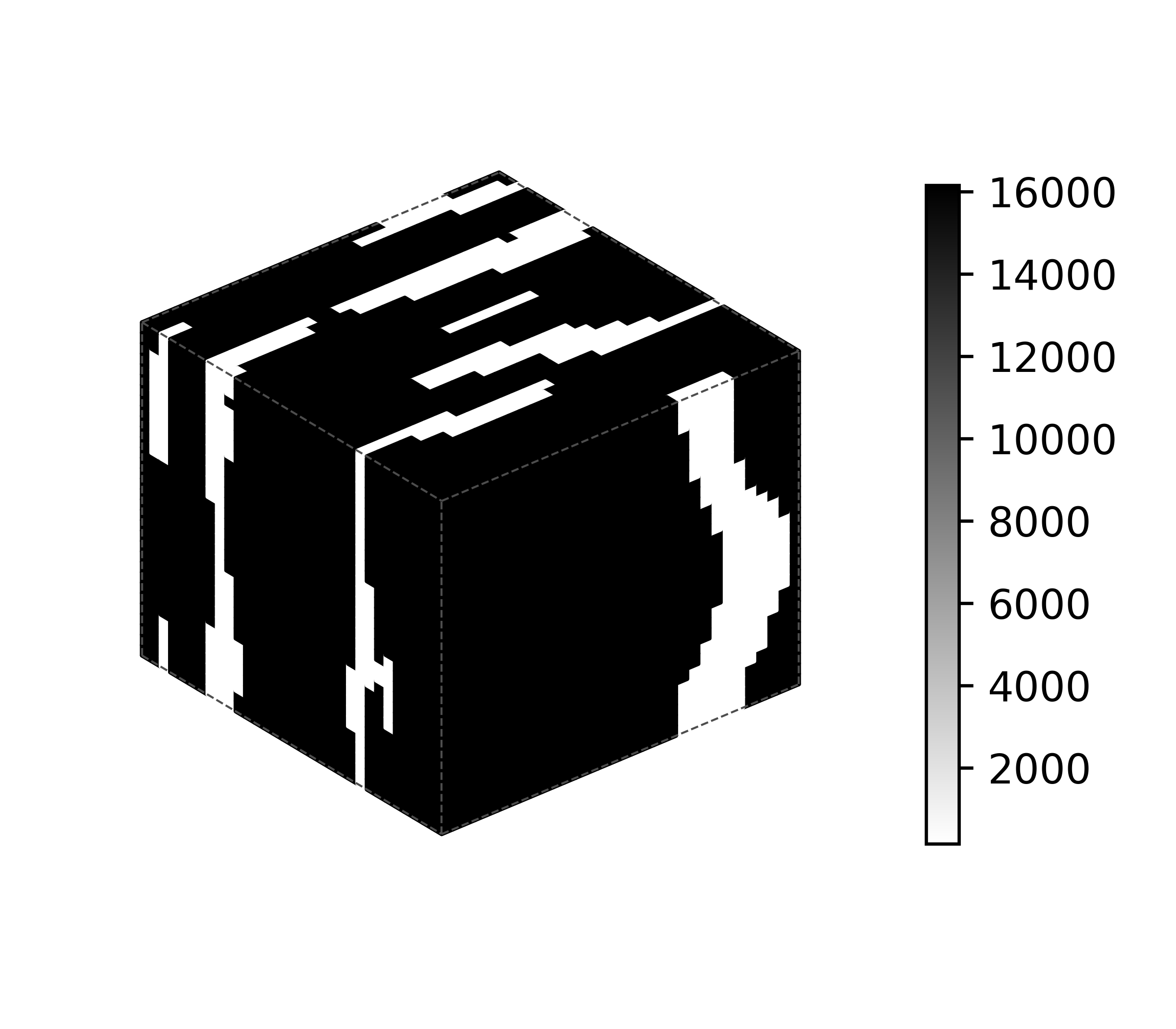}
    \caption{Local $C_{1111}$ stiffness field for a sample microstructure}
    \label{subfig:sample_C}
    \end{subfigure}
    \hfill
    \begin{subfigure}[b]{0.49\textwidth}
    \centering
    \includegraphics[width=\textwidth]{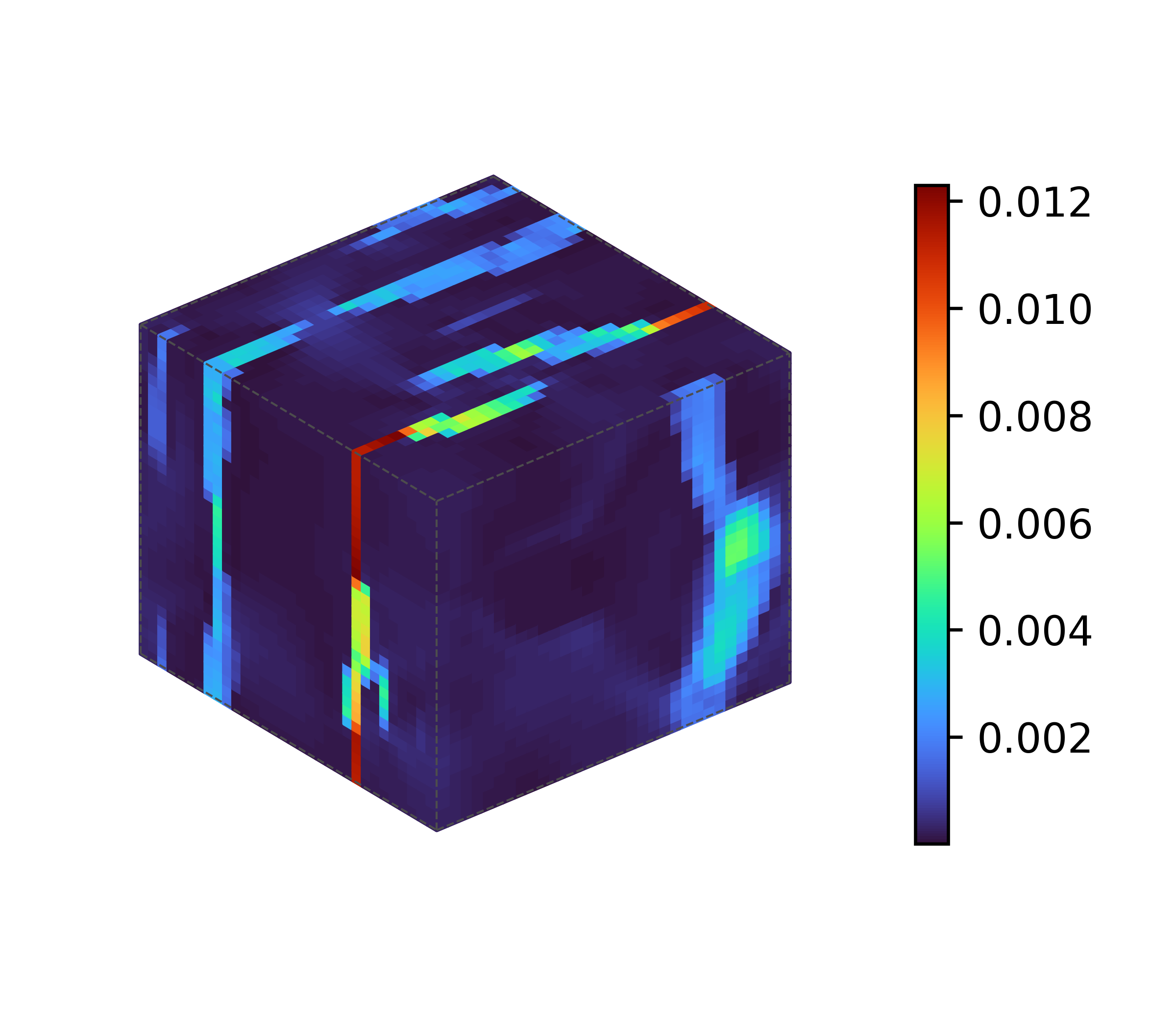}
    \caption{Local equivalent strain of sample}
    \label{subfig:sample_strain}
    \end{subfigure}
    \hfill
    \begin{subfigure}[b]{0.49\textwidth}
    \centering
    \includegraphics[width=\textwidth]{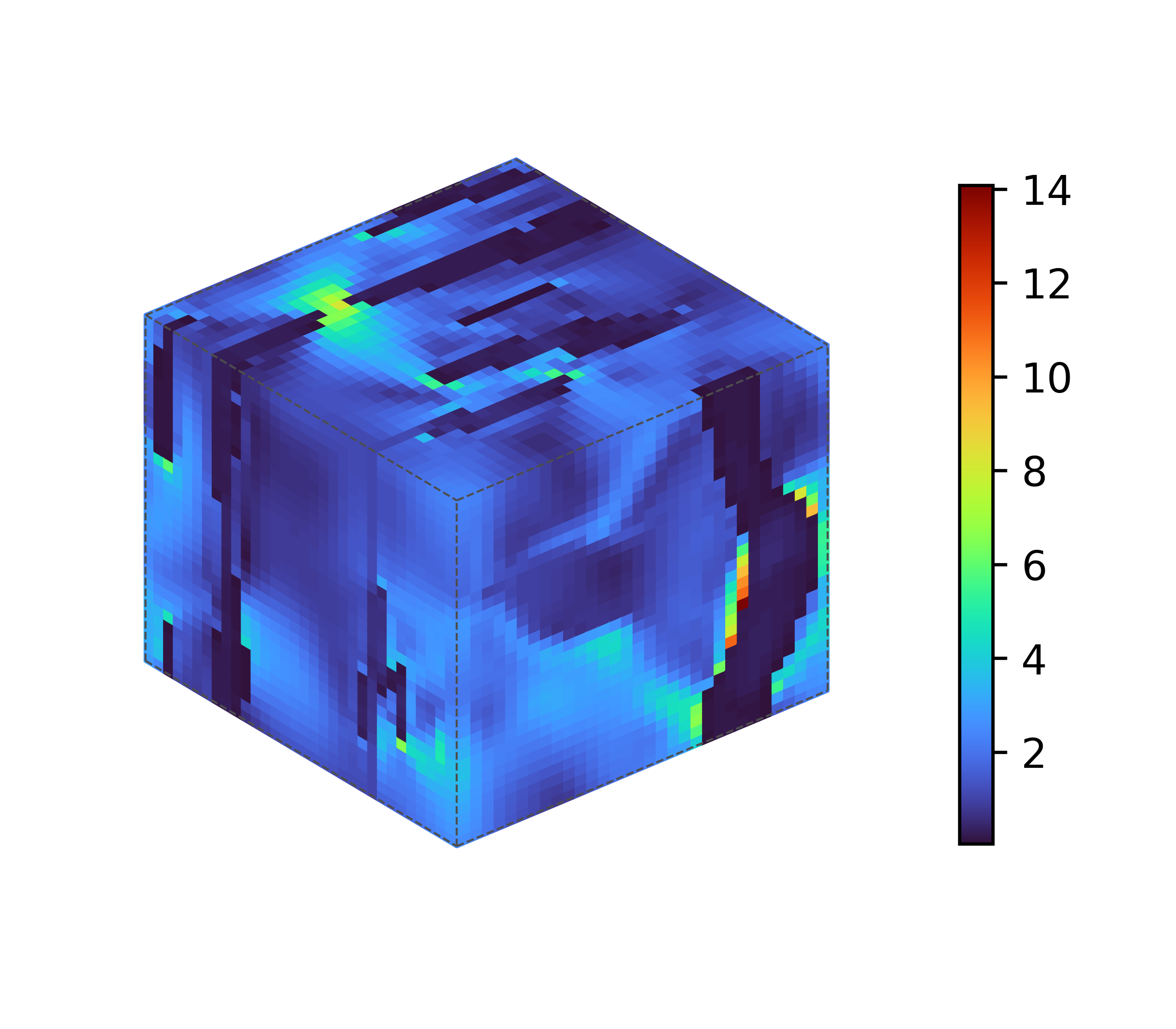}
    \caption{Local equivalent stress of sample }
    \label{subfig:sample_stress}
    \end{subfigure}
    \hfill
    \begin{subfigure}[b]{0.49\textwidth}
    \centering
    \includegraphics[width=\textwidth]{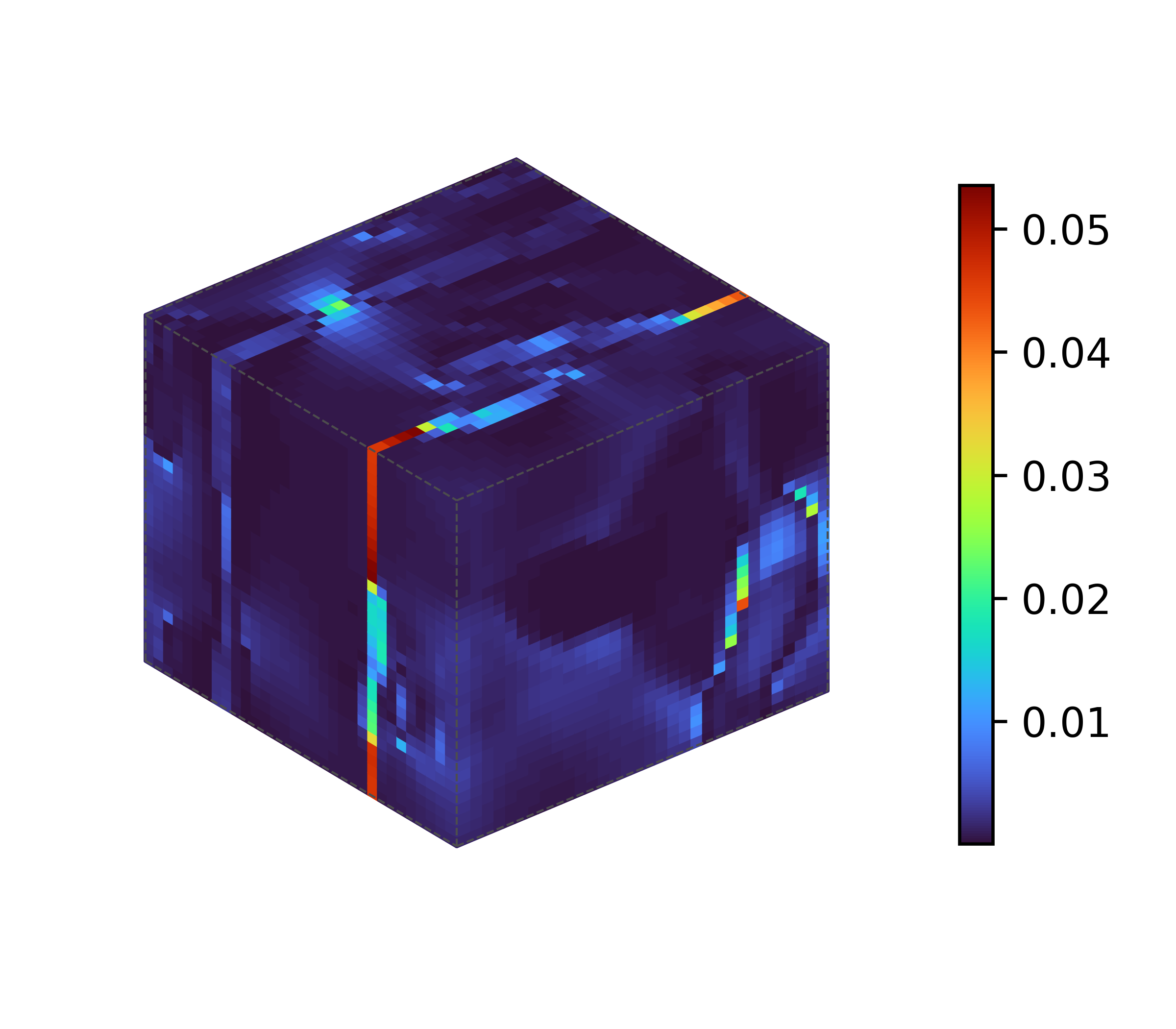}
    \caption{Local strain energy density of sample}
    \label{subfig:sample_energy}
    \end{subfigure}
    \caption{Sample microstructure and solution fields, contrast $\kappa=100$, loading of 0.1\% $x-x$ tensile strain}
    \label{fig:sample}
\end{figure}

Our first case study compares each model with contrast and boundary conditions fixed; all microstructures have contrast $\kappa = 100$ and the loading is always in the $x-x$ direction. This setting was also used for the ablation study and timing comparisons. A representative microstructure (displayed via the $C_{1111}$ field), along with the corresponding strain, stress, and energy fields, is shown in Figure \ref{fig:sample}. A notable feature is the near-percolation of the white (soft) phase, which leads to concentrations of strains and stresses in bottleneck areas.

A table of each trained model's error metrics and average runtime (measured in milliseconds per microstructure) is presented in Table \ref{tab:cs1_results}. These timing results are not necessarily optimal for each architecture (indeed, all of the implementations have significant room for optimization), but provide a helpful, quick comparison between models. 

\begin{table}[h]
    \centering
    \begin{tabular}{@{}l|llll|l@{}}
    \multicolumn{1}{l}{\% Error} & Homog. Stiff. & L1 VM Stress & L2 Strain & L2 Stress & \multicolumn{1}{l}{Avg. inference time (ms)} \\
    \toprule
    FNO & 4.083 & 13.785 & 32.145 & 4.956 & 0.388 \\
    Big FNO & 1.761 & 8.471 & 19.638 & 2.485 & 1.101 \\
    IFNO & 0.734 & 4.603 & 10.848 & 1.545 & 3.568 \\
    FNO-DEQ & 0.742 & 3.841 & 8.922 & 1.198 & 6.549 \\
    \hline
    Mod. F-D & 0.520 & 2.949 & 7.808 & 0.969 & 6.408 \\
    TherINO & \textbf{0.337} & \textbf{2.048} & \textbf{6.316} & \textbf{0.757} & 7.460 \\
    %\bottomrule
    \end{tabular}
\caption{Summary of evaluation metrics for different architectures. All error quantities listed in normalized percentage via the method described in \ref{app:errors}.  Best quantities per category listed in \textbf{bold}.}
\label{tab:cs1_results}
\end{table}

The vanilla FNO acts as a baseline to compare other methods against; it is the fastest of the considered models, but also produces the worst results. During the training and calibration process we found it highly sensitive to changes in hyperparameter and choice of loss function. Increasing the model size provided only a modest performance improvement. The regular FNO achieved 4.08\% accuracy on the homogenized stiffness, compared to 1.76\% for the bigger model.  However, all of the iterative models produced less than 1\% error with the same parameter budget as the baseline FNO. 

Notably, the IFNO was the fastest among the implicit models since it did not require any root-finding procedure. This came with the tradeoff of increased training memory overhead, limiting the resolution and batch size that could be trained on. Comparatively, the FNO-DEQ provided slightly improved error metrics and lower training memory overhead, but was slower due to bookkeeping required for Anderson acceleration.  Modifying the FNO-DEQ to iterate on strain space directly further improved accuracy. This change required more evaluations of the encoder and decoder (a minute slowdown), but also reduced the size of the latent space, reducing the overhead of the fixed-point procedure and leading to a small but stable net speedup. Finally, adding in Thermodynamic Encodings resulted in another jump in accuracy, in exchange for the cost of evaluating the constitutive law every iteration. 

In summary, all implicit models provided significantly more accurate predictions, iterating over strain fields was more effective than iterating over a learned latent space, and the thermodynamic encodings provided the most accurate model by far. We emphasize that even the slowest model required less than 8 milliseconds per microstructure. Each error metric correlated strongly with the others for this case study, but we note that TherINO especially outperformed in the stress-like metrics; the other models only see stress in their loss function, whereas TherINO uses stresses internally during iterations. 

\begin{figure}[h]
    \centering
    \includegraphics[width=\linewidth]{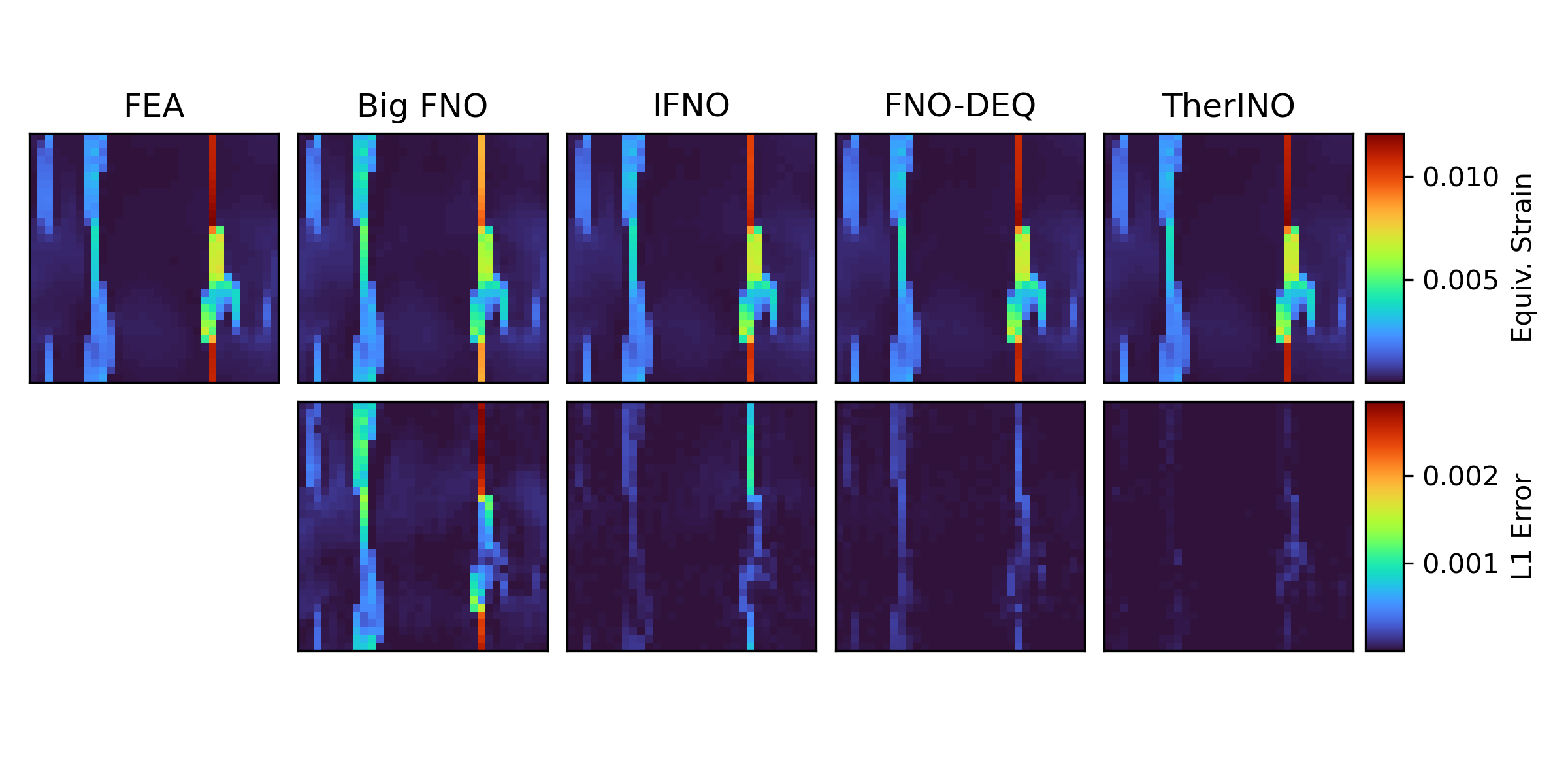}
    \caption{Slices of equivalent strain predictions for fixed-contrast case study. Colorbars are shared across each row; predictions are colored according to the FEA results, and errors are colored according to the extreme error values across all models.}
    \label{fig:cs1_strain}
\end{figure}

\begin{figure}
    \centering
    \includegraphics[width=\linewidth]{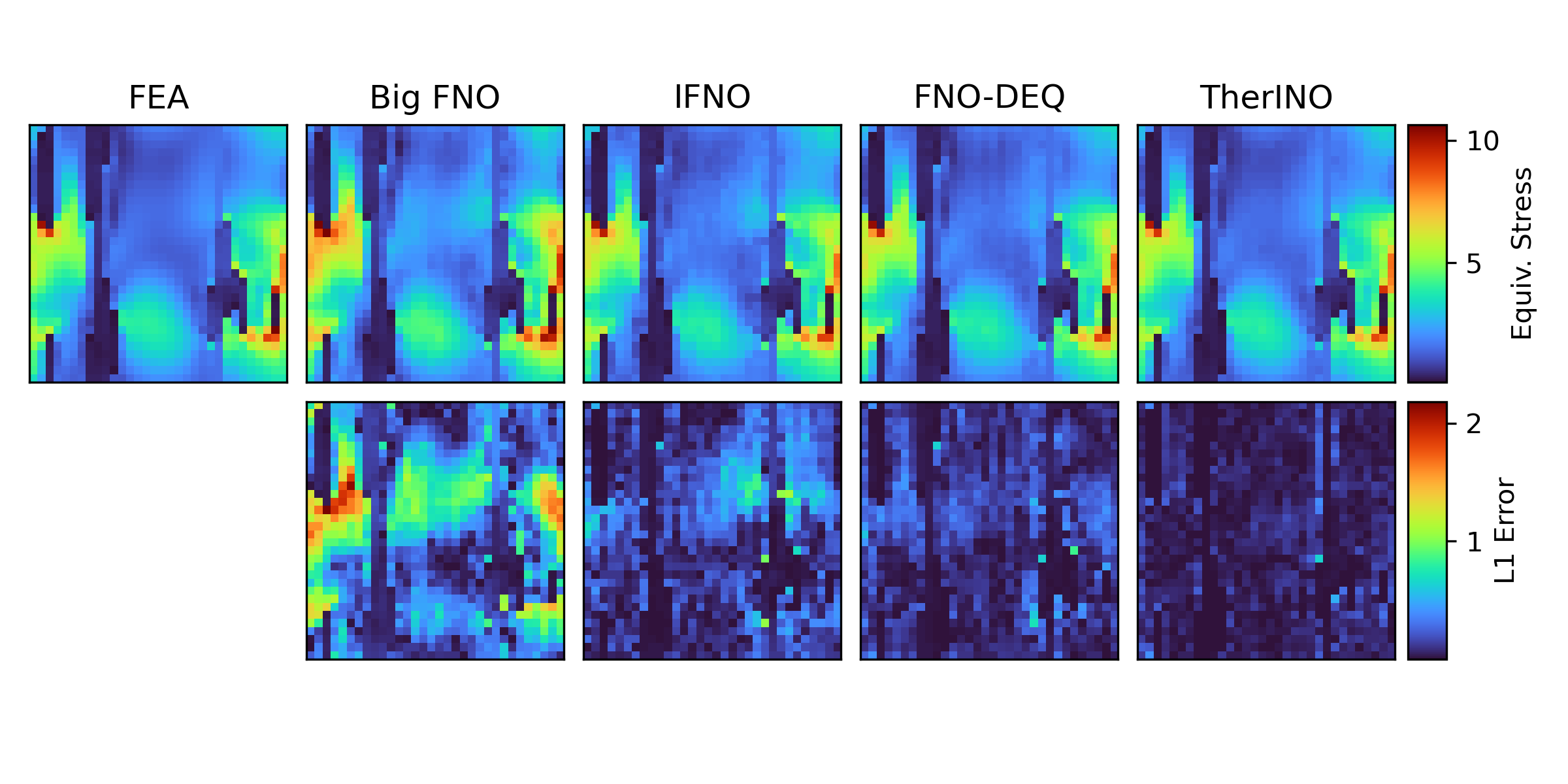}
    \caption{Slices of equivalent stress predictions for fixed-contrast case study. Same colorbar conventions as before}
    \label{fig:cs1_stress}
\end{figure}

In order to provide a qualitative comparison, we plot predictions made for a challenging microstructure in the test set. Since each solution contains $(6 \times 32 \times 32 \times 32)$ pieces of information, we only plot a constant$-z$ slice of equivalent strain and stress fields to simplify the comparison. The results of this procedure are presented in Figures \ref{fig:cs1_strain} and \ref{fig:cs1_stress}, respectively. Visually we see that the FNO struggles with peaks and sharp transitions in the strain field. This agrees with existing literature wherein neural operators suffer a \textit{spectral bias} towards low-frequency information \cite{basri_frequency_nodate, qin_toward_2024, zhang_blending_2024}. Both IFNO and FNO-DEQ handle this peak much better, but still fail to capture the sharp transition. However, TherINO captures the peak value almost perfectly, and its error is spread more evenly through the structure. A similar picture appears in stress space, where we see the FNO creating significant artifacts and overshoots in high-stiffness regimes. Note that the regular FNO and Modified FNO-DEQ are omitted for brevity from this plot; the former produced very poor predictions, visually, and the latter provided no additional information beyond the regular FNO-DEQ. 

\clearpage
\subsection{Randomized boundary conditions}
In this case study, the contrast is fixed to $\kappa = 100$ for all microstructures, and the loading strain $\bceps$ is randomized over the unit $6$-sphere, then scaled to have magnitude $0.001$. This is performed over the same train/test/validation splits as before, and all structures are re-run through the Finite Element solver. Table \ref{tab:cs3_results} shows the same quantitative error metrics as before, except for the homogenized stiffness (whose definition depends on the loading conditions). Since each model's computational cost is independent of contrast and loading direction, timing results are also omitted. Compared to the first case study, all models perform slightly worse, likely due to the higher diversity in response fields.

\begin{table}[h]
    \centering
    \begin{tabular}{@{}l|lll@{}}
    \multicolumn{1}{l}{} & L1 VM Stress & L2 Strain & L2 Stress  \\
    \toprule
    FNO  & 16.010 & 47.468 & 7.278 \\
    Big FNO & 11.048 & 30.924 & 3.995 \\
    IFNO & 4.513 & 13.881 & 2.161 \\
    FNO-DEQ & 3.905 & 12.786 & 1.781 \\
    \hline
    Mod. F-D & 3.704 & 12.945 & 1.544 \\
    TherINO & \textbf{2.582} & \textbf{10.140} & \textbf{1.223} \\
    %\bottomrule
    \end{tabular}
\caption{Evaluation metrics for varying-boundary-condition case}
\label{tab:cs2_results}
\end{table}

While TherINO provides only a modest improvement in local strain predictions, it strongly outperforms the other implicit models in both VM and L2 stress errors. This is reflected clearly in the qualitative comparison; while the equivalent strain field is visually quite similar to the previous study, the equivalent stresses build up in very different parts of the microstructure. Since the other models are not aware of stresses outside of their loss functions, we hypothesize that the inclusion of thermodynamic encodings improves stress predictions \textit{without negatively impacting} strain predictions. Based on these results, it appears that the combination of adding bulk strains as inputs and enforcing average strains produces models which interpolate well for periodic boundary conditions. We note that increasing the \textit{magnitude} of loading strains will have no performance impact as all strain-like quantities are normalized by $\|\bceps\|$. 

\begin{figure}[h]
    \centering
    \includegraphics[width=\linewidth]{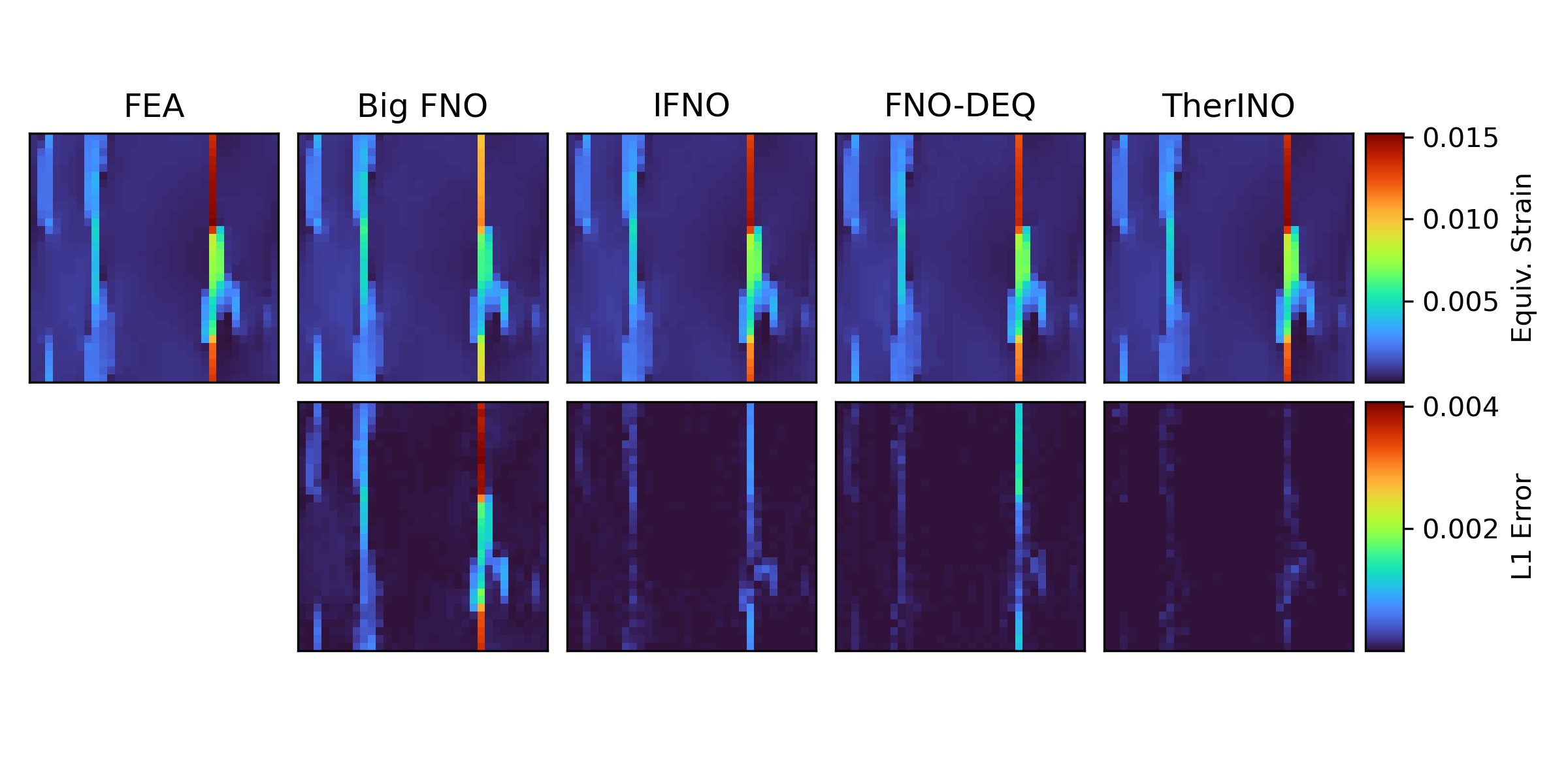}
    \caption{Slices of equivalent strain predictions for variable-boundary-condition case study.}
    \label{fig:cs2_strain}
\end{figure}

\begin{figure}
    \centering
    \includegraphics[width=\linewidth]{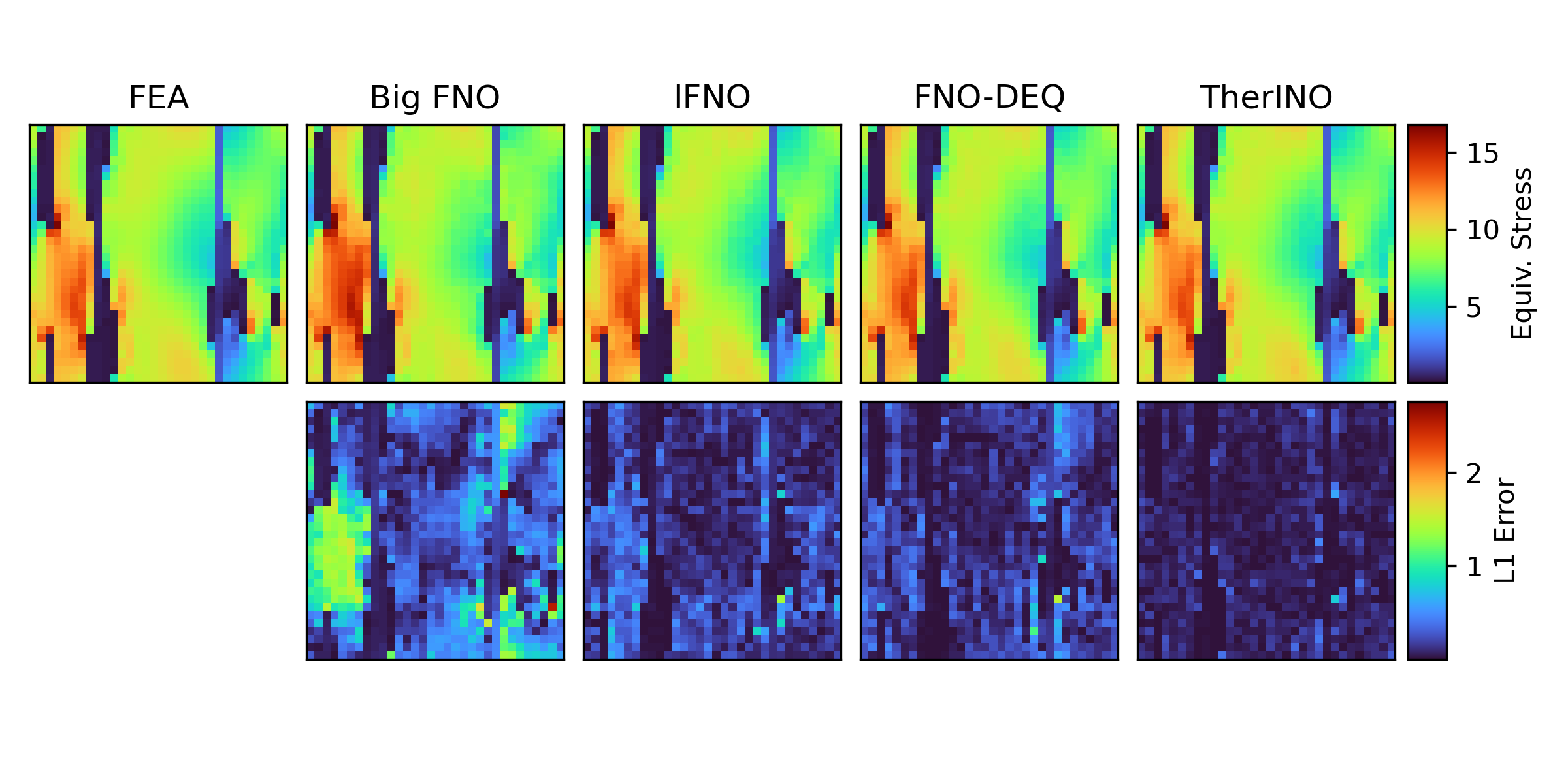}
    \caption{Slices of equivalent stress predictions for variable-boundary-condition case study.}
    \label{fig:cs2_stress}
\end{figure}

\clearpage
\subsection{Randomized contrast}
To evaluate each model's interpolation ability in contrast space, we randomly select $\kappa \in \{2, 10, 50, 100\}$ for each microstructure (again preserving splits).  The same set of error metrics is presented in Table \ref{tab:cs2_results}. 

\begin{table}[h]
    \centering
    \begin{tabular}{@{}l|llll@{}}
    \multicolumn{1}{l}{} & Homog. Stiff. & L1 VM Stress & L2 Strain & L2 Stress  \\
    \toprule
    FNO & 2.106 & 8.055 & 18.213 & 1.983 \\
    Big FNO & 0.747 & 4.292 & 10.487 & 1.063 \\
    IFNO & 0.484 & 2.409 & 5.689 & 0.613 \\
    FNO-DEQ & 0.516 & 2.400 & 5.952 & 0.585 \\
    \hline
    Mod. F-D & 0.521 & 2.545 & 6.808 & 0.587 \\
    TherINO & \textbf{0.300} & \textbf{1.501} & \textbf{4.370} & \textbf{0.360} \\
    %\bottomrule
    \end{tabular}
\caption{Evaluation metrics for variable-contrast dataset.}
\label{tab:cs3_results}
\end{table}

In a sense this is an easier dataset than the first two: $\kappa$ is upper-bounded by the first study and the loading direction is fixed. This is reflected by a decrease in testing error across the board. The same general trend of performance increases carries over, with the notable exception that the modified FNO-DEQ slightly \textbf{worse} than the regular FNO-DEQ. However, the inclusion of thermodynamic encodings still significantly outperforms the comparison models. In other words, strain-space iteration allows for more flexible models, but does not seem strictly better on its own. We omit the qualitative comparison for this problem as is visually very similar to the first case study.

Since the entire training set was for contrast $\kappa \leq 100$, one naturally questions how well the model extrapolates \textit{outside} the training distribution. To examine this we take the model trained on variable contrasts, and apply it to the test set with an increased contrast $\kappa = 200$ fixed across all microstructures. In other words, \textit{how well does the model extrapolate in stiffness space}? In order to preserve stability, we recalculate the reference stiffness and scaling (e.g. normalize relative to the new contrast). We note that this calculation depends only on $\kappa$, which is required to compute local stiffness fields. This scaling therefore preserves the sanctity of the test set, but avoids numerical instability due to extrapolation in stiffness space.

Table \ref{tab:cs3.5_results} presents the results of this extrapolation study; as expected, the accuracy of all models degrades significantly. Note that these models are identical -- weight for weight -- with those in Table \ref{tab:cs3_results}, except for the stiffness scaling values. While the modified FNO-DEQ produced better VM stresses and homogenized stiffnesses than learned iterations, it had comparable or worse local strain predictions. TherINO suffered a significantly smaller degradation in performance, especially in homogenized stiffness where its error was almost 5 times lower than IFNO and 2.5 times lower than the modified FNO-DEQ.

\begin{table}[h]
    \centering
    \begin{tabular}{@{}l|llll@{}}
    \multicolumn{1}{l}{} & Homog. Stiff. & L1 VM Stress & L2 Strain & L2 Stress  \\
    \toprule
    FNO & 13.378 & 24.487 & 44.228 & 5.945  \\
    Big FNO & 10.661 & 17.099 & 32.199 & 3.539  \\
    IFNO & 9.867 & 12.784 & 21.622 & 2.335  \\
    FNO-DEQ & 13.515 & 18.311 & 19.828 & 2.207  \\
    \hline
    Mod. F-D & 5.097 & 10.285 & 21.401 & 2.008  \\
    TherINO & \textbf{2.009} & \textbf{5.083} & \textbf{13.881} & \textbf{1.146}  \\
    %\bottomrule
    \end{tabular}
\caption{Evaluation metrics for out-of-distribution contrast test.}
\label{tab:cs3.5_results}
\end{table}

\begin{figure}
    \centering
    \includegraphics[width=\linewidth]{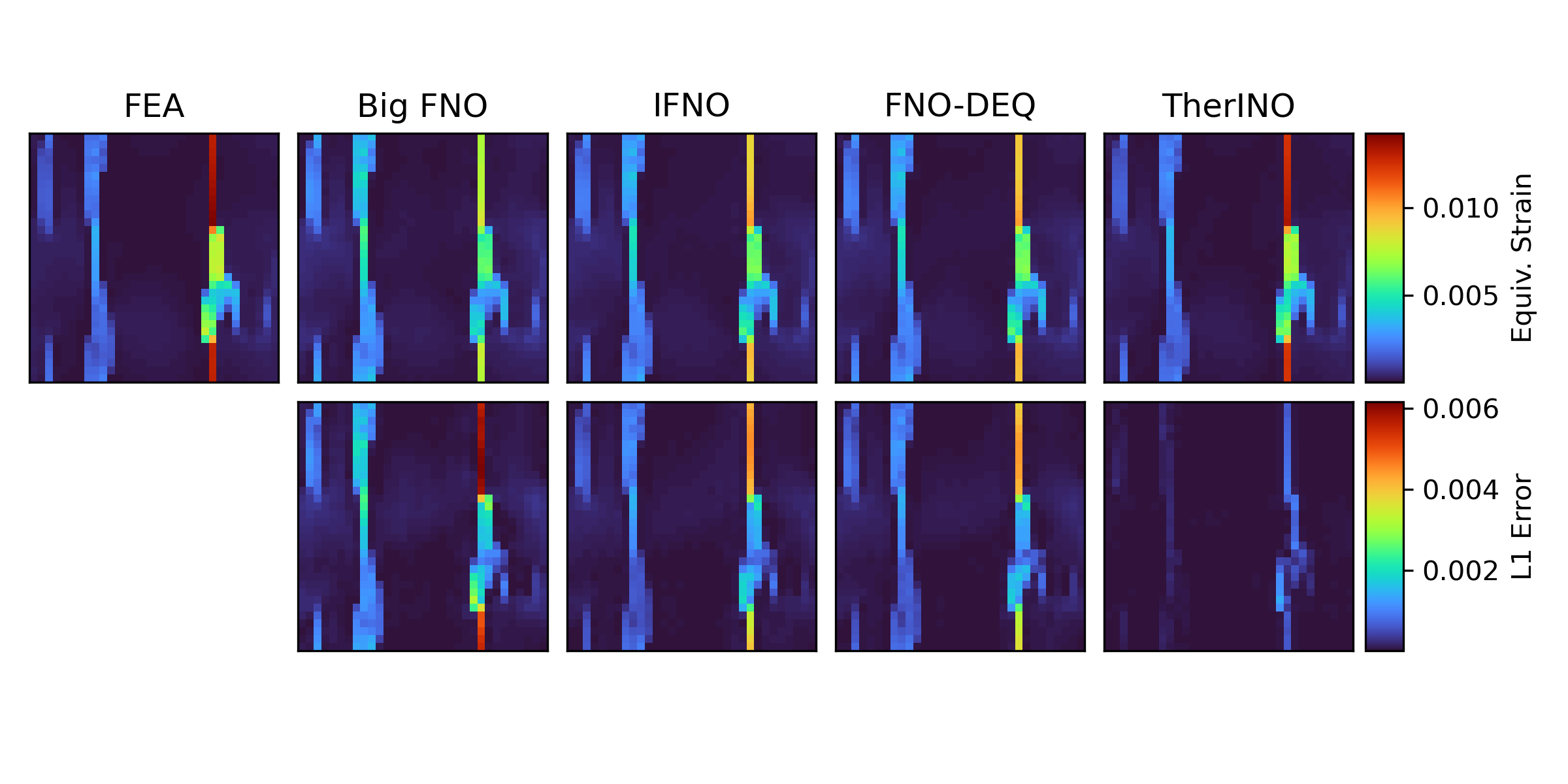}
    \caption{Slices of equivalent strain predictions for contrast-extrapolation case study.}
    \label{fig:cs3.5_strain}
\end{figure}

\begin{figure}
    \centering
    \includegraphics[width=\linewidth]{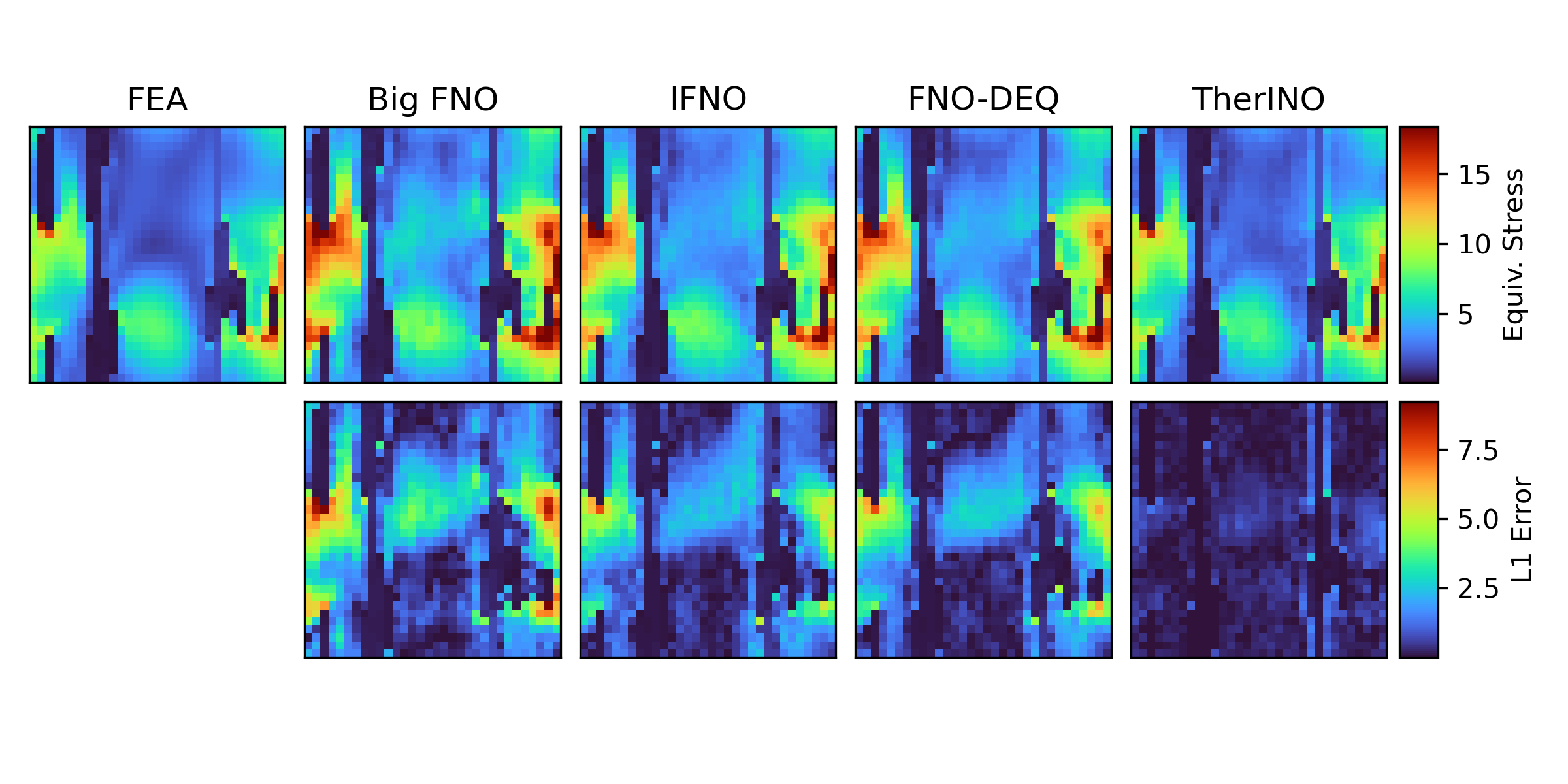}
    \caption{Slices of equivalent stress predictions for contrast-extrapolation case study. Same colorbar conventions as before.}
    \label{fig:cs3.5_stress}
\end{figure}

Again we plot slices of strain and stress predictions for each model on the extrapolated dataset in Figures \ref{fig:cs3.5_strain} and \ref{fig:cs3.5_stress}, respectively. Especially notable is the sharp central stripe -- whereas IFNO and FNO-DEQ previously captured the stripe well, they fail to capture the magnitude of the peak in the high-contrast case. Conversely, TherINO is relatively stable. In stress space the same artifacts are increased in severity for models which did not use thermodynamic encodings. 

In summary, the combination of strain-space iteration and thermodynamic encodings provides better accuracy both in- and out-of-distribution. Strain-space iteration alone has some advantages, but is not inherently better on its own. 

\FloatBarrier

\subsection{DEQ Convergence}
\begin{figure}
    \centering
    \includegraphics[width=\linewidth]{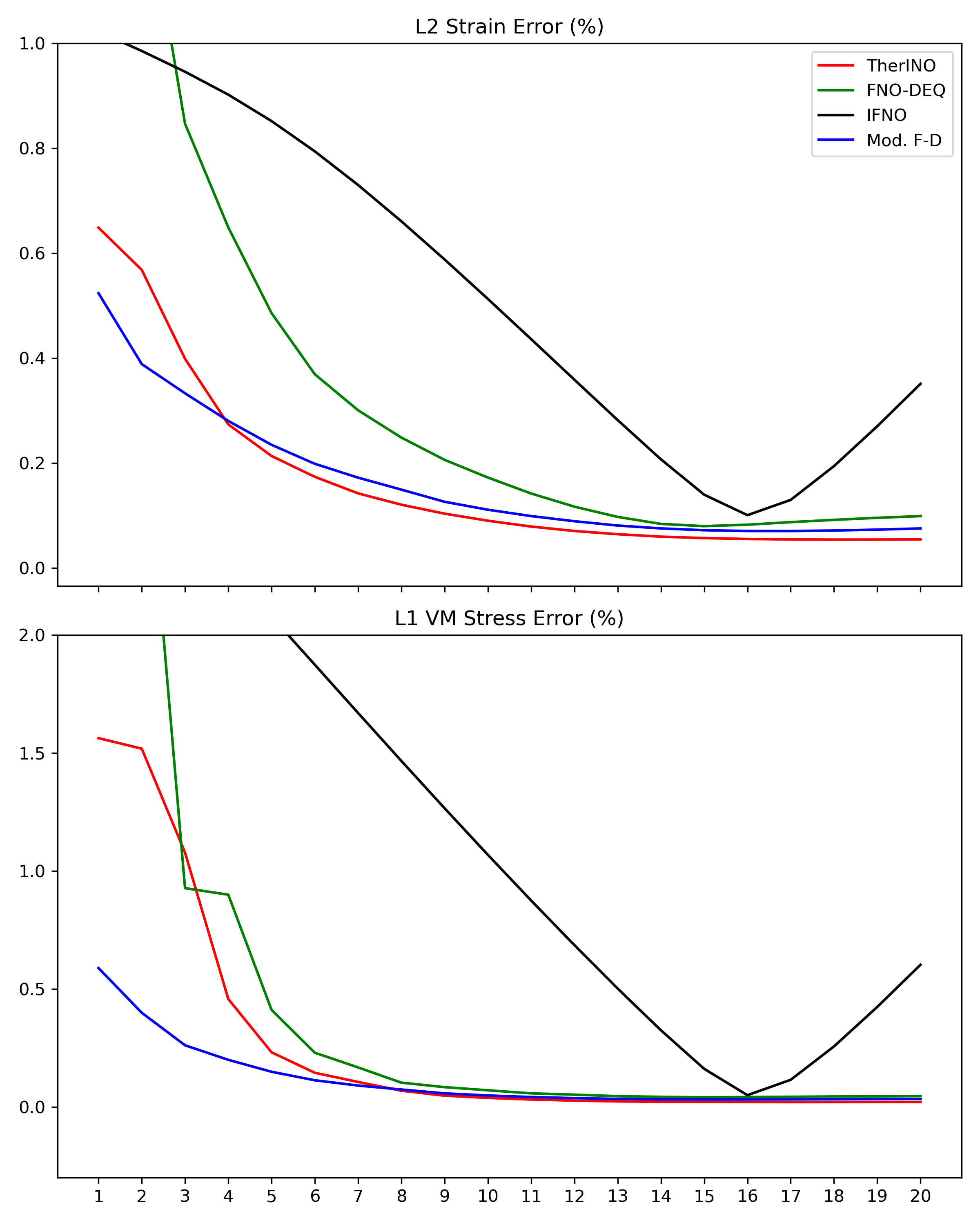}
    \caption{L2 strain and stress errors averaged over 100 randomly-selected test set microstructures. All models were trained on the dataset from Case Study 1.}
    \label{fig:conv_errors}
\end{figure}

To close our analysis, we also explore the rate of convergence of each iterative model towards its converged solution, and how error changes across iterations. Since our architecture does not guarantee contractivity, the DEQ models are neither guaranteed to have a fixed point nor to converge to one if it exists.  Moreover, it is not immediately clear what effect the thermodynamic encodings will have on numerical stability during DEQ rollout. While the empirical results above indicate that these are not severe limitations, we wish to elucidate the rate of convergence in both DEQ residuals and solution trajectories. To explore stability beyond the training horizon (16 iterations), we evaluate each model for a total of 20 iterations. To check these convergence properties, Figure \ref{fig:conv_errors} presents the L2 strain and stress errors (averaged over 100 test-set structures) across iterations. For comparison the IFNO trajectory is also plotted; note that since it learns an ODE flow rather than a fixed point, it is unable to stop at the desired solution and instead overshoots after the training cutoff. In contrast, all three DEQ models seem to converge to a stable prediction. However, the error seems to plateau rather than decreasing with more iterations. This means that we could potentially obtain a speedup by stopping iterations sooner (e.g., trade off inference accuracy for inference time), but also provides an upper bound on the model accuracy. 

\begin{figure}
    \centering
    \includegraphics[width=\linewidth]{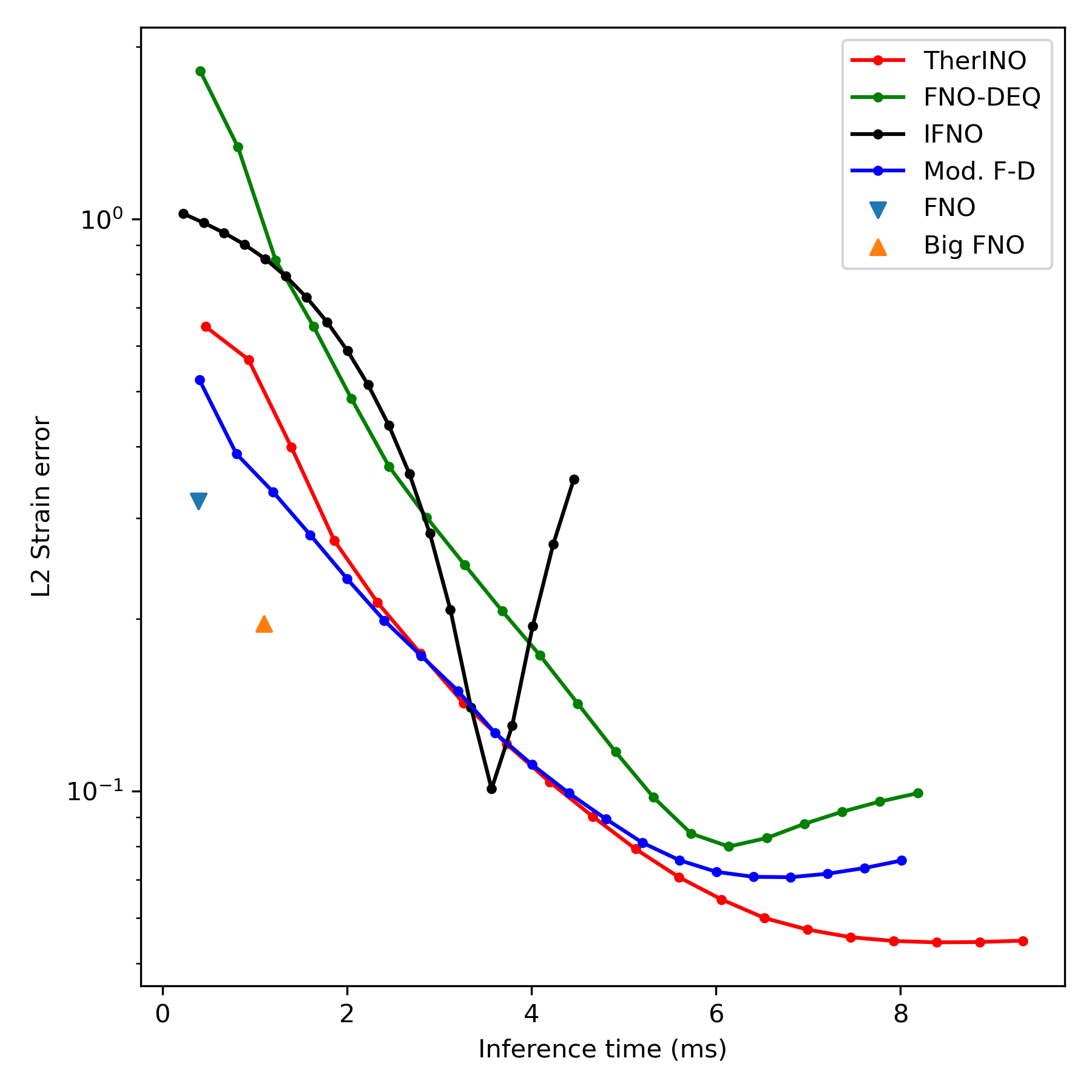}
    \caption{Accuracy-speed tradeoff for each method on the fixed-contrast dataset}
    \label{fig:tradeoff}
\end{figure}

Plotting the strain results \textit{vs.} inference time (computed from Table \ref{tab:cs1_results}) for each model produces Figure \ref{fig:tradeoff}. We observe that while the FNO models are fastest for low-accuracy predictions, they are quickly overtaken by the iterative neural operators. Due to its speed, IFNO is able to quickly make a medium-accuracy prediction, but cannot perform any further refinement. Among the DEQ-based models, the modified FNO-DEQ performed better in the low-accuracy regime (likely due to faster initial convergence), but TherINO caught up and outperformed it in later iterations. Notably, the regular FNO-DEQ performed worst among the iterative models in this plot -- clearly the faster and stronger fixed-point convergence did not equate with improved prediction accuracy.

\FloatBarrier

To examine stability for DEQ models we also display a normalized DEQ residual (using $\mathbf{h}_i$ to represent either a latent state or a candidate strain, depending on model) across iterations:

\begin{equation}
    r_i = \frac{\|f_\phi(\mathbf{h}_i) - \mathbf{h}_i \|}{\| \mathbf{h}_i \|}.
\end{equation}
\noindent We also plot the alignment $a_i$ between search direction $s_i$ and error $e_i$ in strain space for each iteration:
\begin{align}
    \mathbf{s}_i &= Q(\mathbf{h}_{i+1}) - Q(\mathbf{h}_i) \\
    \mathbf{e}_i &= \beps - Q(\mathbf{h}_i) \\
    a_i &= \frac{\mathbf{s}_i \ddp \mathbf{e}_i}{\sqrt{\mathbf{s}_i \ddp \mathbf{s}_i} \sqrt{\mathbf{e}_i \ddp \mathbf{e}_i}}.
\end{align}

\begin{figure}
    \centering
    \includegraphics[width=\linewidth]{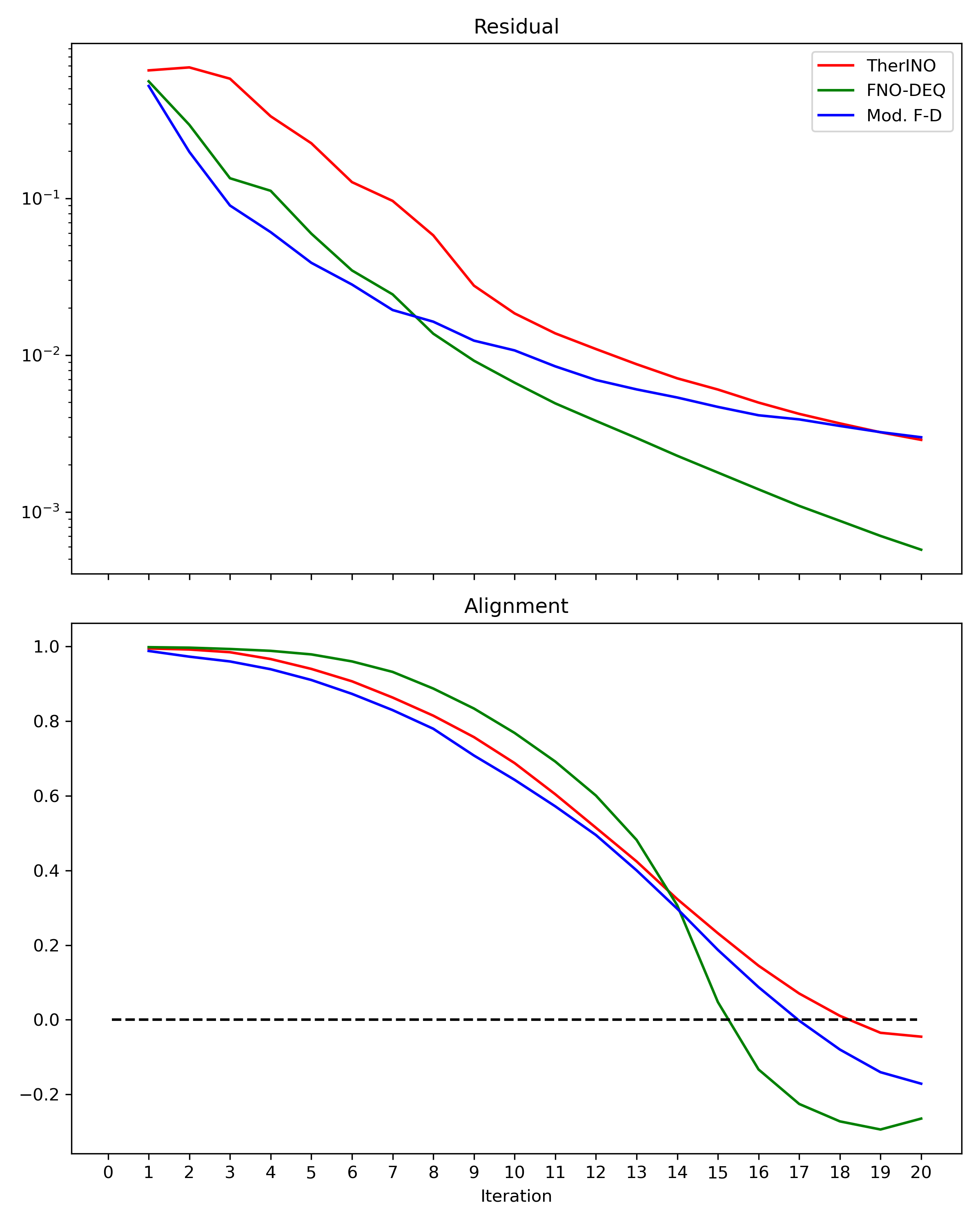}
    \caption{DEQ residual and alignment for the same trajectories as previous.}
    \label{fig:conv_resids}
\end{figure}

These traces are presented in Figure \ref{fig:conv_resids}. It is immediately apparent that the strain-space iteration hinders the DEQ's convergence -- both the Modified FNO-DEQ and TherINO show stagnation in their residual, whereas the basic FNO-DEQ converges roughly linearly. In other words, it appears that learning a latent space allows the DEQ to find a more contractive mapping, in exchange for worse accuracy and extrapolability. This is reinforced by the alignment trace, which shows that the full DEQ significantly overshoots the true strain field. In other words, the inexact fixed-point solve encourages each model to find a ``good enough'' answer within 16 iterations, rather than as a true fixed point of the model. 

\FloatBarrier
\section{Conclusions}

In this work we developed and benchmarked a class of implicit neural operators which use constitutive information as model inputs. By testing a variety of material contrasts and loading directions, we showed that thermodynamic encodings provide flexibility and generalizability atypical for most machine learning methods. The central challenge of applying data-driven techniques to materials science, particularly at the mesoscale, is \textit{data scarcity}. By definition, materials discovery requires making predictions outside the space of known microstructures. Therefore any learning model must be at least somewhat robust to distribution shifts. In this respect our model provides a major advancement; for the task of zero-shot extrapolation to higher-contrast microstructures, TherINO provided the most accurate answers and suffered the smallest generalization gap. 

However, these improvements come with the drawback of a massive hyperparameter space. There are innumerable design choices, but slim literature for the design of implcit neural operators. To address this, we have also provided a number of analytical tools and discussion on the training of implicit neural operators for 3D heterogeneous localization. To the best of the author's knowledge, this is the first work to apply a neural operator to heterogeneous elastic localization in the 3D materials domain. While we use elasticity to develop the model, our work also provides a pathway towards modeling more complex, time-dependent, and irreversible phenomena. 

This work presents a number of directions for future analysis: first, the extension beyond linear elasticity. For hyperelastic materials, this could be done by simply replacing the linear strain-to-stress map with a nonlinear constitutive equation. We expect that the thermodynamic encodings will offer even larger benefits as the dependence of solution on coefficients becomes increasingly nonlinear. Extension to large-deformation systems would be somewhat more involved, as it would require replacing the strain and stress measures appropriately, and converting boundary conditions to the correct reference frame. Second, the iteration over strain space theoretically allows our model to be blended with a traditional solver. This direction has been explored somewhat in literature via the HINTS method, but in that context a relaxation solver was the driver (with a neural operator acting as a kind of preconditioner). Our model could theoretically also be used as a preconditioner, but could also use a relaxation solver as part of the thermodynamic encodings. Third, the current architecture is restricted to a regular grid on a cubic domain; this could ostensibly be alleviated by usage of a fixed coordinate embedding between a regular grid and the domain of interest, or by changing the underlying filtering operation to a DeepONet or Graph Neural operator. Fourth, it would be informative to explore the impact of underlying neural operator on our model's contractivity and stability, as well as techniques to guarantee existence and stability of fixed points. Finally, our proposed model did not guarantee compatibility or equilibrium in any function space, but this could be imposed by incorporating weak-form PDE residuals (e.g. \cite{goswami_physics-informed_2022,rezaei_finite_2024, harandi_spectral-based_2024} and predicting deformation fields, then differentiating appropriately to obtain compatible strains. 

\section{Code and Data Availability}
All software used for data generation, model training, and evaluation will be provided under an open-source license. All model weights datasets will be freely available online following publication. 

\section{CRediT authorship contribution statement}
\textbf{Conlain Kelly}: Conceptualization, Methodology, Software, Data curation, Investigation, Formal analysis, Writing – original draft, Writing – review \& editing. \textbf{Surya R. Kalidindi}: Supervision, Funding acquisition, Writing – original draft, Writing – review \& editing.

\section{Acknowledgments}
C. Kelly would like to thank Andreas Robertson, Rafael Orozco, and Ziyi Yin for numerous conversations in the development of this work, as well as Sébastien Brisard for his guidance regarding FFT solvers. C. Kelly acknowledges support from NSF 2027105, NSF GRFP DGE-1650044, Sandia LDRD Contract Agreement \#2177713. S.R. Kalidindi and C. Kelly acknowledge support from ONR N00014-18-1-2879. This work utilized computing resources and services provided by the Partnership for an Advanced Computing Environment (PACE) and the Hive computing cluster at the Georgia Institute of Technology, Atlanta, Georgia, USA.

\clearpage
\appendix
\section{Notation and data representation}\label{app:representation}

Table \ref{tab:notation} provides definitions for common notation in this work
\newcolumntype{L}{>{$}l<{$}} % math-mode version of "l" column type

\begin{table}[h]
    \centering
    \begin{tabular}{L|l}
        \text{Variable} &  Definition \\ \hline
        % M & Stochastic Microstructure Function \\
        m(x) & Microstructure  \\ % (sampled from M) \\ 
        \bC(x) & Elastic stiffness tensor field \\
        \beps(x) & Generic strain field \\
        \bsig(x) & Generic stress field \\
        w(x) & Generic strain energy density field \\
        \bceps & Imposed (reference) elastic strain field \\
        \trueeps & ``True'' solution strain field \\
        \candeps & Predicted strain field  \\
        \volavg{\cdot} & Volume-average operation \\
        \phi & Neural network weights and tunable parameters
    \end{tabular}
    \caption{Definitions of common notation in this work}
    \label{tab:notation}
\end{table}

We compute scaling coefficients for strain and stiffness independently of the training set (using Frobenius norms of the average strain and reference stiffness, respectively), and scale downstream variables appropriately:

\begin{align}
    \beps_{scaled} &= \frac{\beps}{\|\bceps\|_2} \\
    \bC_{scaled} &= \frac{\bC}{\|\bC^0\|_F} \\
    \bsig_{scaled} &= \frac{\bsig}{\|\bC^0\|_F \|\bceps\|_2}  \\
    w_{scaled} &= \frac{w}{\|\bC^0\|_F^2 \|\bceps\|_2}.
\end{align}

All quantities are scaled before being input into any models, and unscaled after being output. All figures are unscaled (e.g. in original units), but loss and error terms are computed using scaled quantities. 

All strain- and stress-like quantities are represented internally using Mandel notation. This representation vectorizes a symmetric $3 \times 3$ matrix $\bsig$ as follows:

\begin{equation}
    vec(\bsig) = \begin{bmatrix}
        \sigma_{11} \\
        \sigma_{22} \\ 
        \sigma_{33} \\
        \sqrt{2} \sigma_{23} \\
        \sqrt{2} \sigma_{13} \\
        \sqrt{2} \sigma_{12}
    \end{bmatrix}.
\end{equation}

Likewise, a (minor-) symmetric rank-4 $\bC$ tensor of shape $3 \times 3 \times 3 \times 3$ is matricized:

\begin{equation}
    mat(\bC) = \begin{bmatrix}
    C_{1111} & C_{1122} & C_{1133} & \sqrt{2}C_{1123} & \sqrt{2}C_{1113} & \sqrt{2}C_{1112} \\
    C_{2211} & C_{2222} & C_{2233} & \sqrt{2}C_{2223} & \sqrt{2}C_{2213} & \sqrt{2}C_{2212} \\
    C_{3311} & C_{3322} & C_{3333} & \sqrt{2}C_{3323} & \sqrt{2}C_{3313} & \sqrt{2}C_{3312} \\ \sqrt{2}C_{2311} & \sqrt{2}C_{2322} & \sqrt{2}C_{2333} & 2 C_{2323} & 2 C_{2313} & 2 C_{2312} \\ \sqrt{2}C_{1311} & \sqrt{2}C_{1322} & \sqrt{2}C_{1333} & 2 C_{1323} & 2 C_{1313} & 2 C_{1312} \\ \sqrt{2}C_{1211} & \sqrt{2}C_{1222} & \sqrt{2}C_{1233} & 2 C_{1223} & 2 C_{1213} & 2 C_{1212}
    \end{bmatrix}.
\end{equation}

Note that these operations preserve inner product and norms; that is, for rank-2 tensors $\bsig, \beps$ and rank-4 tensor $\bC$, we have

\begin{align}
    vec(\bsig)^T \cdot vec(\beps) = \bsig \ddp \beps \\
    mat(\bC) \cdot vec(\mathbf{\beps}) = vec(\bC \ddp \beps)  \\
    mat(\bC) \cdot mat(\bC) = mat(\bC \ddp \bC)
\end{align}

where $(\cdot)$ represents standard matrix-matrix and matrix-vector multiplication. 

\section{Architecture Design and ablation}\label{app:arch}

For all models we use input-injected FNO blocks with no mode truncation (e.g., preserve all frequency information in the microstructure). For all models we used 2 FNO layers (e.g. Fourier-convolution, linear transform, and activation), with injection occurring before the activation. Except for the Big FNO, the internal channel width was fixed to 24 throughout the FNO's body. The Adam optimizer (with standard settings) was used in combination with a Cosine Annealing learning rate decay, from an initial value of $5e-3$ down to a final value of $1e-6$ over 200 epochs. The initial weights of all Fourier convolutional filters were chosen uniformly in the range $[0, \alpha]$, where $\alpha$ is a hyperparameter representing the initial weight scale. The choice of $\alpha = 0.1$ allowed the iterative models to train stably and had minimal performance impact.

Inputs to the lifting block are normalized (using GroupNorm with one group to emulate LayerNorm) and are lifted using a pointwise linear transform (i.e., a $1 \times 1 \times 1$ ``local'' convolution). As an alternative we considered a pointwise MLP (two local convolutions with an activation in between), but this led to unstable training in some situations without improving accuracy. The projection block is implemented as a pointwise MLP through a larger channel space -- in this case, moving through a 128-channel intermediate proved essential to get sharp local predictions. We found that weight normalization on all local convolutions improved training stability, and GELU activations were used throughout. For comparison we also tried ReLU and PReLU activations; the former trained somewhat slower and converged to worse models, and the latter drastically increased GPU memory pressure due to the need to store intermediate activations to tune channel weights. During training we used all-zeros as an initial state for the FNO-DEQ and the constant loading strain as an initial state for the modified FNO-DEQ and TherINO. During evaluation we used all-zero initial states for each model, although this choice did not seem to make a strong impact on convergence or accuracy. 

For all DEQ-based models, we used a maximum of 16 iterations for backwards passes during training. Inspired by the results of \cite{anil_path_2022}, the number of forward passes was randomized between 2 and 30 (imposing an average of 16) during training and fixed to 16 during evaluation. We found that this added some noise to the training process, but drastically improved the overall training stability and model performance. We also used gradient clipping (to unit norm) for all models; this helped the iterative models avoid divergence early in training. DEQ models were trained using IFT gradients, rather than phantom gradients or backpropogation-through-time \cite{fung_jfb_2022}; we experimented with the aforementioned alternatives, but encountered convergence issues (possibly due to the inclusion of thermodynamic encodings). 

To confirm our choices of hyperparameter, we conducted a grid search over the FNO width and depth, with the results presented in Table \ref{tab:ablation_arch}. Note that while the larger models performed better, they required \textbf{significantly} more tunable parameters and showed signs of overfit. Both model types had a performance gap between train and validation sets, but the gap widened after 3 layers, especially for TherINO. Empirically, we found similar performance characteristics on $32^3$ datasets and for the IFNO and FNO-DEQ models, so we chose a depth of 2 and width of 24 for all iterative methods. The last row (depth-4, width-48) is the `Big FNO' used in the case studies above. All of the architecture search was conducted with a maximum learning rate of $5e-3$ and a weight scale of $1$, except for the width-48 models which used a weight scale of $0.1$ for stability.

\begin{table}[h]
    \centering
    \begin{tabular}{@{}l|rrrrrr|rrr@{}}
        \multicolumn{1}{l}{} &  Homog. &  VM &  Strain &  Stress &  Train loss &  \multicolumn{1}{l}{Valid. loss} &  Width &  Depth & Params \\
        \toprule
        \multirow{7}{*}{FNO}  &          7.434 &         23.08 &               40.42 &               7.295 &        2.080e-1 &        1.924e-1 &                        24 &               1 &     1185020 \\
        &           4.14 &         13.05 &               27.56 &               4.683 &        8.073e-2 &        9.382e-2 &                        16 &               2 &     1052668 \\
        &          3.222 &         11.91 &               26.57 &              \textbf{ 4.296 }&        6.027e-2 &        8.846e-2 &                        24 &               2 &     2365340 \\
        &          3.012 &         11.46 &               26.46 &               4.157 &        4.631e-2 &        9.164e-2 &                        32 &               2 &     4202556 \\
        &          1.957 &         9.801 &               22.59 &               3.371 &        1.748e-2 &        6.709e-2 &                        24 &               3 &     3545660 \\
        &          1.523 &         8.385 &               18.31 &               2.654 &        6.009e-3 &        4.557e-2 &                        24 &               4 &     4725980 \\
        &          \textbf{1.488} &      \textbf{   7.709 }&              \textbf{ 17.25} &              \textbf{ 2.354} &       \textbf{ 1.413e-3} &       \textbf{ 4.160e-2} &                        48 &               4 &    18892796 \\
        \hline
        \multirow{7}{*}{TherINO}  &         0.7805 &          3.98 &                11.3 &               1.336 &        1.370e-2 &        1.606e-2 &                        24 &               1 &     1184812 \\
        &         0.4448 &         2.602 &               8.072 &              0.9897 &        6.600e-3 &        8.162e-3 &                        16 &               2 &     1052524 \\
        &         0.3348 &         2.273 &               6.804 &              0.8838 &        3.711e-3 &        6.134e-3 &                        24 &               2 &     2365132 \\
        &        \textbf{ 0.2714} &         2.189 &               6.367 &              0.8328 &        2.441e-3 &        5.425e-3 &                        32 &               2 &     4202284 \\
        &          0.334 &         \textbf{2.077} &               \textbf{ 5.77} &              \textbf{0.8321} &        1.811e-3 &        \textbf{4.850e-3} &                        24 &               3 &     3545452 \\
        &          0.306 &         2.264 &               5.996 &              0.8939 &        1.303e-3 &        5.293e-3 &                        24 &               4 &     4725772 \\
        &         0.4555 &         2.314 &               6.807 &               0.878 &        \textbf{4.736e-4} &        6.852e-3 &                        48 &               4 &    18892396 \\
    %\bottomrule
    \end{tabular}
    \caption{Ablation study for varying architectures}
    \label{tab:ablation_arch}
\end{table}

Keeping the weight and depth fixed to $24$ and $2$, respectively, we next explored the space of initial learning rate (affecting training stability and convergence) and initial weight scaling (affecting model stability and local optima). The results of this search are presented in Table \ref{tab:ablation_hyper}; an initial learning rate of $0.005$ led to fastest and most stable training for in all cases. The initial weight scaling proved less important, leading only to slightly different local optima for each model. However, for the iterative models using too large of initial weight scaling led to poor convergence occasionally, so we chose an initial scale of $0.1$ for all FNO layers. 

\begin{table}[h]
    \centering
    \begin{tabular}{@{}l|rrrrrr|rr@{}}
        \multicolumn{1}{l}{} &  Homog. &  VM &  Strain &  Stress &  Train loss &  \multicolumn{1}{l}{Valid. loss} &  Max LR &  Weight Scale \\
        \toprule
        \multirow{5}{*}{FNO} &           3.36 &         12.03 &               26.91 &               4.308 &        5.881e-2 &        9.234e-2 &   0.005 &                         0.1 \\
        &          3.371 &         12.19 &               26.37 &               4.419 &        6.302e-2 &        8.781e-2 &   0.001 &                           1 \\
        &          \textbf{3.222} &        \textbf{ 11.91} &               26.57 &               4.296 &        6.027e-2 &        8.846e-2 &   0.005 &                           1 \\
        &          3.312 &         12.06 &               26.85 &               4.386 &        5.926e-2 &        9.126e-2 &    0.01 &                           1 \\
        &          3.385 &         12.08 &               \textbf{26.28} &               4.351 &        \textbf{5.854e-2} &       \textbf{ 8.736e-2} &   0.005 &                          10 \\
        \hline
        \multirow{5}{*}{TherINO} &        \textbf{ 0.2929} &        2.355 &               6.836 &              0.9038 &        \textbf{3.581e-3} &        \textbf{6.124e-3} &   0.005 &                         0.1 \\
        &         0.4746 &         2.781 &               8.396 &              0.9946 &        6.605e-3 &        8.985e-3 &   0.001 &                           1 \\
        &         0.3348 &         \textbf{2.273} &              \textbf{ 6.804} &            \textbf{  0.8838} &        3.711e-3 &        6.134e-3 &   0.005 &                           1 \\
        &         0.3344 &         2.573 &               7.314 &              0.9295 &        3.976e-3 &        6.708e-3 &    0.01 &                           1 \\
        &         0.3176 &         2.419 &               7.352 &              0.9279 &        4.082e-3 &        7.484e-3 &   0.005 &                          10 \\
        %\bottomrule
    \end{tabular}
    \caption{Ablation study for varying hyperparameters}
    \label{tab:ablation_hyper}
\end{table}

\FloatBarrier
`
\section{Training data generation}\label{app:train_data}

All synthetic 2-phase microstructures were generated using a filtered Gaussian Random Field. The design parameters for these synthetic microstructures are the characteristic feature sizes in each direction and the relative volume fraction of each phase, as well as the Young's Modulus and Poisson ratio of each phase. A Latin Hypercube Sampling strategy was used to sample the feature sizes and volume fraction, after which a synthetic microstructure was produced by sampling Gaussian noise, filtering, and thresholding. We note that during preparation of this work a number of new methods for synthetic microstructure generation have arisen, and would be an interesting extension to this work. 

To obtain stress and strain solution fields we used the Moose Finite Element Solver. For all results we use a mesh matching the microstructure resolution ($32^3$) as our ``ground truth''. While not strictly necessary for the following analysis, this drastically simplifies the generation pipeline and lowers the wall time required to construct a synthetic dataset. This does allow the dual risks of bleeding FEA artifacts into the neural operator predictions and allowing the FNO to learn aliasing artifacts due to under-resolved training data. However, we remark that super-resolution of solution fields is less important than \textit{accurately predicting quantities of interest}; in the future we plan to apply this method to experimental data, which is inherently noisy. 

Within the solver we used trilinear hexahedral elements will full integration. Rather than work with nodal displacements, we extract stress and strain values at element centroids and use those as piecewise-constant voxel fields. To compute solutions, a stabilized biconjugate gradient (BiCGSTAB) linear solver was used with a multigrid preconditioner (BoomerAMG) \cite{henson_boomeramg_2002}. While mesh refinement and more sophisticated averaging would produce higher-quality predictions, the goal here is to see how well a neural operator can match even medium-fidelity FEA simulations, and how that predictive capacity affects prediction of downstream quantities of interest. 

All models are constrained to have the correct average strains by first subtracting off the instance-wise average strains, then adding back in $\bceps$. In the second case study we explore each model's ability to interpolate within this space. Since each microstructure in this dataset had the same contrast, we could have equivalently used $\{0, 1\}$ phase-wise indicators instead of the full $21$ stiffness components. However, empirically there was no difference in training or testing performance between these two representations for any architecture considered, so we chose the most general representation. This flexibility is leveraged heavily in the third case study.

\section{Error metric calculation} \label{app:errors}
To calculate error metrics, we first calculate error quantities instance-wise, then average over the test set. Effective $x-x$ stiffnesses are calculated (where applicable) using the ratio of average $x-x$ stress to average $x-x$ strain (fixed to $0.001$ in this work) for a given stiffness-strain pair:

\begin{equation}
    C_{eff}(\bC, \varepsilon) = \frac{\volavg{\bC \ddp \beps}_{11}}{\varepsilon_{11}}.
\end{equation}

\noindent The error in homogenized stiffness is computed as the absolute error in $C_{1111}^*$ of the FEA and predicted strains, normalized by the FEA homogenized stiffness, averaged over all instances:

\begin{equation}
    err_{homog} = \sum_i \frac{|C_{eff}(\bC_i, \trueeps_i) - C_{eff}(\bC_i, \candeps_i)|}{C_{eff}(\bC_i, \trueeps_i)}.
\end{equation}

To compute equivalent strains and stresses, we first introduce the deviator of a symmetric tensor $\mathbf{\tau}$ as an orthogonal projection onto the space of traceless (deviatoric) tensors:
\begin{align}
    \tau^{dev}_{ij} &= \tau_{ij} - \frac{1}{3}\tau_{kk} \delta_{ij}. \\ \intertext{Equivalent stresses are therefore defined as }
    \sigma^{equiv} &= \sqrt{\frac{3}{2} \bsig^{dev} \ddp \bsig^{dev}} \\
    \intertext{and the work-conjugate equivalent strains as }
    \beps^{equiv} &= \sqrt{\frac{2}{3} \beps^{dev} \ddp \beps^{dev}} .
\end{align}
\noindent The von Mises (or equivalent) stress error is then computed by calculating FEA and predicted stresses $\sigma_i^*$ and $\hat{\sigma}_i$, computing the relative mean absolute error in their equivalent components, and again averaging over the test set:

\begin{equation}
    err_{VM} = \sum_i \frac{\volavg{|{\sigma_i^*}^{equiv} - \hat{\sigma_i}^{equiv}|}}{\volavg{|{\sigma_i^*}^{equiv}|}}
\end{equation}

Finally, L2 strain and stress errors are computed simply by taking square-rooting the scaled squared-L2 error terms \eqref{eq:loss_strain} and \eqref{eq:loss_stress} \textit{before} averaging instance-wise. This corresponds to an average L2 function norm of the error across the whole dataset.

\clearpage
\bibliographystyle{elsarticle-num}
\bibliography{references}

\end{document}